\documentclass[10pt,journal,compsoc]{IEEEtran}
\usepackage{cite}
\usepackage{algorithmic}
\usepackage{array}
\usepackage{amssymb}
\usepackage{eqparbox}
\usepackage{url}
\usepackage{multirow}
\usepackage[cmex10]{amsmath}
\usepackage{makeidx}
\usepackage{diagbox}
\usepackage{rotating,soul}
\usepackage{wrapfig}
\usepackage{booktabs}
\usepackage[usenames,dvipsnames]{color}
\usepackage{ragged2e}
\usepackage{epsfig}
\usepackage{caption}
\usepackage{overpic}
\usepackage{enumitem}
\usepackage{color}
\usepackage[dvipsnames,table,xcdraw]{xcolor}
\usepackage[linesnumbered,ruled]{algorithm2e}
\usepackage{colortbl}
\usepackage{graphicx}
\usepackage{contour}
\usepackage{listings}

\makeindex

\DeclareGraphicsExtensions{.pdf,.jpg,.png}

\usepackage[pdfborder=false,colorlinks,bookmarksnumbered,bookmarksopen,linkcolor=black,citecolor=black,urlcolor=black]{hyperref}

\DeclareGraphicsExtensions{.jpg,.pdf,.png}

\usepackage{array}
\newcolumntype{x}[1]{>{\centering\arraybackslash\hspace{0pt}}p{#1}}

\usepackage{pifont}

\definecolor{mygreen}{RGB}{0,100,0}
\definecolor{myblue}{RGB}{10,100,200}
\definecolor{myred}{RGB}{200,0,0}

\newcommand{\addFig}[1]{}
\newcommand{\addFigs}[1]{}

\definecolor{mygreen}{RGB}{0,150,0}
\definecolor{myred}{RGB}{200,0,0}

\begin{document}


\title{MADAv2: Advanced Multi-Anchor Based Active Domain Adaptation Segmentation}


\author{Munan Ning, Donghuan Lu, Yujia Xie, Dongdong Chen, Dong Wei, Yefeng Zheng, \\Yonghong Tian, Shuicheng Yan, Li Yuan
  \IEEEcompsocitemizethanks{
    \IEEEcompsocthanksitem Munan Ning, Yonghong Tian, Li Yuan are with Peking University, School of Electronic and Computer Engineering, Shenzhen Graduate School, China, and also with PengCheng Laboratory.  E-mail:\{munanning, yhtian, yuanli-ece\}@pku.edu.cn.
    \IEEEcompsocthanksitem Donghuan Lu, Dong Wei, Yefeng Zheng are with Tencent Jarvis Lab, Shenzhen, China.
	\IEEEcompsocthanksitem Yujia Xie, Dongdong Chen are with the Microsoft Cloud AI, Redmond, WA 98052 USA.
    \IEEEcompsocthanksitem Shuicheng Yan is with Sea AI Lab.  E-mail:yansc@sea.com.
	}
}

\markboth{IEEE TRANSACTIONS ON PATTERN ANALYSIS AND MACHINE INTELLIGENCE,~Vol.~xx,No.~xx,~xxx.~xxxx}%
{Ning \MakeLowercase{\emph{et al.}}: MADAv2: Advanced Multi-Anchor Based Active Domain Adaptation Segmentation}

\IEEEcompsoctitleabstractindextext{%
\begin{abstract}
\justifying
Unsupervised domain adaption has been widely adopted in tasks with scarce annotated data. 
Unfortunately, mapping the target-domain distribution to the source-domain unconditionally may distort the essential structural information of the target-domain data, leading to inferior performance.
To address this issue, we firstly propose to introduce active sample selection to assist domain adaptation regarding the semantic segmentation task. 
By innovatively adopting multiple anchors instead of a single centroid, both source and target domains can be better characterized as multimodal distributions, in which way more complementary and informative samples are selected from the target domain.
With only a little workload to manually annotate these active samples, the distortion of the target-domain distribution can be effectively alleviated, achieving a large performance gain.
In addition, a powerful semi-supervised domain adaptation strategy is proposed to alleviate the long-tail distribution problem and further improve the segmentation performance.
Extensive experiments are conducted on public datasets, and the results demonstrate that the proposed approach outperforms state-of-the-art methods by large margins and achieves similar performance to the fully-supervised upperbound, \emph{i.e.}, 71.4\% mIoU on GTA5 and 71.8\% mIoU on SYNTHIA.
The effectiveness of each component is also verified by thorough ablation studies. 
Code is available at \url{https://github.com/munanning/MADAv2}.
\end{abstract}

}

\maketitle
\IEEEdisplaynontitleabstractindextext
\IEEEpeerreviewmaketitle

\IEEEraisesectionheading{\section{Introduction}\label{sec:introduction}}



\IEEEPARstart{A}{s} a fundamental task of computer vision, image segmentation has been long studied. In the recent decade, the rapid development of deep learning has brought great advances to the various tasks on top of image segmentation, such as autonomous driving~\cite{geiger2012we}, scene parsing~\cite{cordts2016cityscapes,Piao_2019_DMRA}, object detection~\cite{Ji_2021_DCF,Zhang_2019_MoLF,li2021joint,ji2022promoting} and human-computer interaction~\cite{oberweger2015hands}.
However, it also leads to a hunger for huge training data, which is usually laborious and costly to obtain, especially for some expertise-demanding or complicated applications, \emph{e.g.}, medical image segmentation~\cite{Ji_2021_MRNet,ning2020macro,ning2021ensembled,ning2021new,ma2021abdomenct} and auto-driving tasks~\cite{choi2020cars}. 
This data insufficiency issue has greatly limited the application of automatic image segmentation in real-world scenarios.

\begin{figure*}[!ht]
	\centering
	\includegraphics[width=1.8\columnwidth]{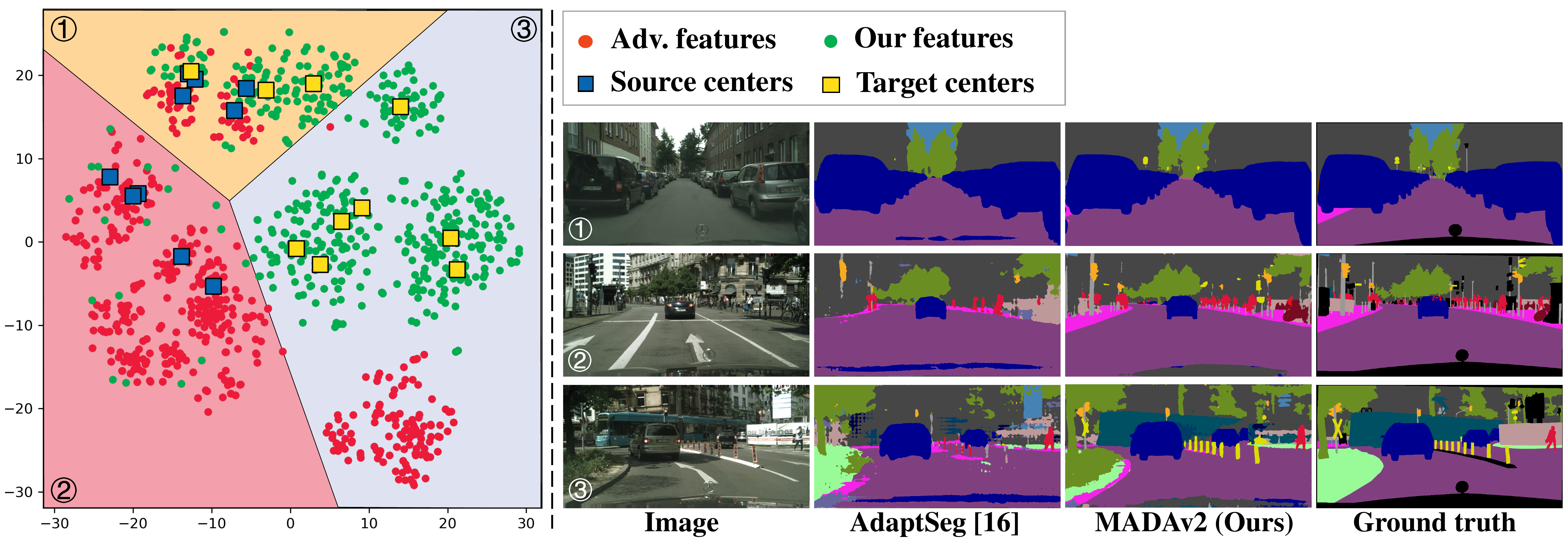}
	\caption{Visualization (t-SNE~\cite{van2008visualizing}) of the target-domain distribution distortion problem in UDA.
Left: Visualization of feature scatters by AdaptSeg~\cite{tsai2018learning} and MADAv2. The source cluster centers (blue squares) and the target cluster centers (yellow squares) can be very close, as shown in region \ding{172}, meaning samples here share similar content; the feature clusters may also show a source- and target-specific distribution, \emph{i.e.},  containing only source or target centers, as shown in region \ding{173} and region \ding{174}, respectively.
The features of AdaptSeg (red dots) are dragged away from the target centers, forced to align with the source centers (see region \ding{173}) or even lose alignment (see region \ding{174}).
In contrast, our features are perfectly distributed around the target centers.
Right: Exemplar samples and corresponding segmentation from AdaptSeg and our method MADAv2. Column 1 shows three samples respectively from regions \ding{172}, \ding{173} and \ding{174}; Columns 2 and 3 are segmentation results with AdaptSeg and our method, respectively. It can be seen that, AdaptSeg works well for the sample from regions \ding{172} as the target samples in region \ding{172} are similar to source samples, while for samples from region \ding{173} and \ding{174}, its segmentation is error-prone as the features do not follow the target centers. 
In contrast, our method can generate precise segmentation for all the samples.
From these examples, we can see that the features of adversarial training tend to deviate from the real target distribution.}
	\label{fig:cluster_visual_1}
\end{figure*}

One of the representative paradigms to solve this issue is unsupervised domain adaptation (UDA)~\cite{chen2017no,hoffman2018cycada,hoffman2016fcns,tsai2018learning}.
UDA methods try to align the target-domain distribution towards the source-domain distribution, and apply the networks trained with the supervision of only the source data to the target data. 
Though the UDA methods have gained impressive achievements, they tend to undermine the latent structural pattern of the target domain as a result of them forcing representations of the target-domain to fit the distribution of the source-domain, which may lead to substantial performance degradation.

We provide in Fig.~\ref{fig:cluster_visual_1} an illustration of the distribution distortion of the target-domain features caused by applying UDA methods, \emph{e.g.}, AdaptSeg \cite{tsai2018learning}, which is a typical adversarial training based UDA method.
Fig.~\ref{fig:cluster_visual_1} (left) is the t-SNE~\cite{van2008visualizing} visualization of the latent representations. 
After applying AdaptSeg \cite{tsai2018learning}, most adapted target-domain features (red dots) are dragged away from target centers (yellow squares) and forced to align with the source centers (blue squares), as shown in region \ding{173}, or lose alignment as shown in region \ding{174}. 
We also show three exemplar images respectively from regions \ding{172}, \ding{173} and \ding{174} with their corresponding segmentation by AdaptSeg in Fig.~\ref{fig:cluster_visual_1} (right).
It can be seen that AdaptSeg only performs well in region \ding{172} where source and target samples share similar content, and generates error-prone segmentations in regions \ding{173} and \ding{174}. 
As demonstrated in Fig.~\ref{fig:cluster_visual_1}, adversarial UDA methods tend to cause the generated target features to deviate from the real distribution (\emph{i.e.}, to be distorted), thus losing some target-specific information and leading to bad segmentation.

To alleviate the above distribution distortion problem, a promising idea is to introduce the knowledge from the real target distribution, \emph{i.e.}, selecting and annotating target samples and learning target-specific information from the annotated sample pairs. 
This can be done with active learning (AL) \cite{settles2009active}, which aims at getting better performance via a little annotation workload, with effectiveness well proved in the domain adaptation (DA) scenery for classification and detection tasks~\cite{su2020active}.
However, most previous AL methods~\cite{su2020active} select samples only based on the target-domain distribution, which may not be optimal to training the network jointly with source-domain data as in common practice and lead to inferior performance.

Based on these observations, in our preliminary conference paper, we proposed a Multi-anchor Active Domain Adaptation (MADA) framework~\cite{ning2021multi}, which adopts the active learning strategy to assist DA regarding the semantic segmentation task, with a multi-anchor strategy to better characterize the source-domain and target-domain features.
Specifically, the MADA framework consists of two stages. In the first stage, with the network pretrained in an adversarial UDA \cite{tsai2018learning} manner, the most complementary samples are selected through the proposed multi-anchor strategy by exploiting the feature distribution across the source domain. 
Then in the second stage, the segmentation network is fine-tuned in a semi-supervised learning manner. The annotations of the source samples and the few selected target samples are jointly used for supervision, and all the available image information is additionally employed for optimization with a pseudo label loss and the proposed multi-anchor soft-alignment loss. 

Though with noticeable improvement, our MADA method still shows an obvious gap with the fully-supervised upperbound, \emph{i.e.}, 64.9\% vs. 71.9\%, a gap of 7.0\% of mIoU on GTA5~\cite{richter2016playing} and 68.1\% vs. 73.0\%, 4.9\% gap on SYNTHIA~\cite{ros2016synthia}.
We arguably attribute such a gap to the sample selection method and the semi-supervised domain adaptation strategy.
Hence in this work, we substantially extend our previously proposed MADA framework and propose an advanced multi-anchor based active domain adaptation segmentation framework (MADAv2).
For sample selection, our previous MADA~\cite{ning2021multi} adopts a single-domain selection method, which takes the distance between target samples and corresponding source anchors as the only metric, neglecting the possibility that selected samples may be \textit{crowded} in the target domain and fail to provide sufficient knowledge of the whole target domain.
In MADAv2, we argue that the selected samples should not only be complementary to the source domain, but also to most other target samples, such that more information of the target domain can be provided. 
Therefore, we extend the single-domain metric to a \textit{Dual\_Domain\_Distance} active metric, \emph{i.e.}, considering the distances from target samples to both source and target anchors comprehensively.
Specifically, we select those target samples that are far away from the target anchors, which would bring better performance than the close ones (the result of \textit{Dual\_Domain\_Distance} is 65.1\% mIoU on GTA5, better than 62.6\% of the close ones).
The new sample selection metric leads to an improvement of 0.9\% mIoU on GTA5, compared to the 64.2\% result of MADA, as experimentally validated in Section~\ref{sec:sample_selection}.
In addition, the previous compact semi-supervised phase in MADA cannot handle the long-tail distribution problem in the target domain~\cite{hu2021semi}, which is extremely severe due to the limited annotations.
The tail classes are easily ignored with the cross-entropy loss because of their few pixels and the similarity between unlabeled target samples and corresponding pseudo labels.
Therefore in MADAv2, we adopt the online hard example mining (OHEM) loss~\cite{shrivastava2016training}, which is designed to find hard examples in object detection tasks, to locate informative pixels and exploit the inconsistency between prediction and pseudo labels. 
Moreover, a combination of revised adaptive cutmix and copy-paste~\cite{hu2021semi} is introduced as additional data augmentation to force the network to pay more attention to tail classes and alleviate the long-tail distribution problem.

\begin{figure}[ht]
	\centering
	\small
	\includegraphics[width=1.0\columnwidth]{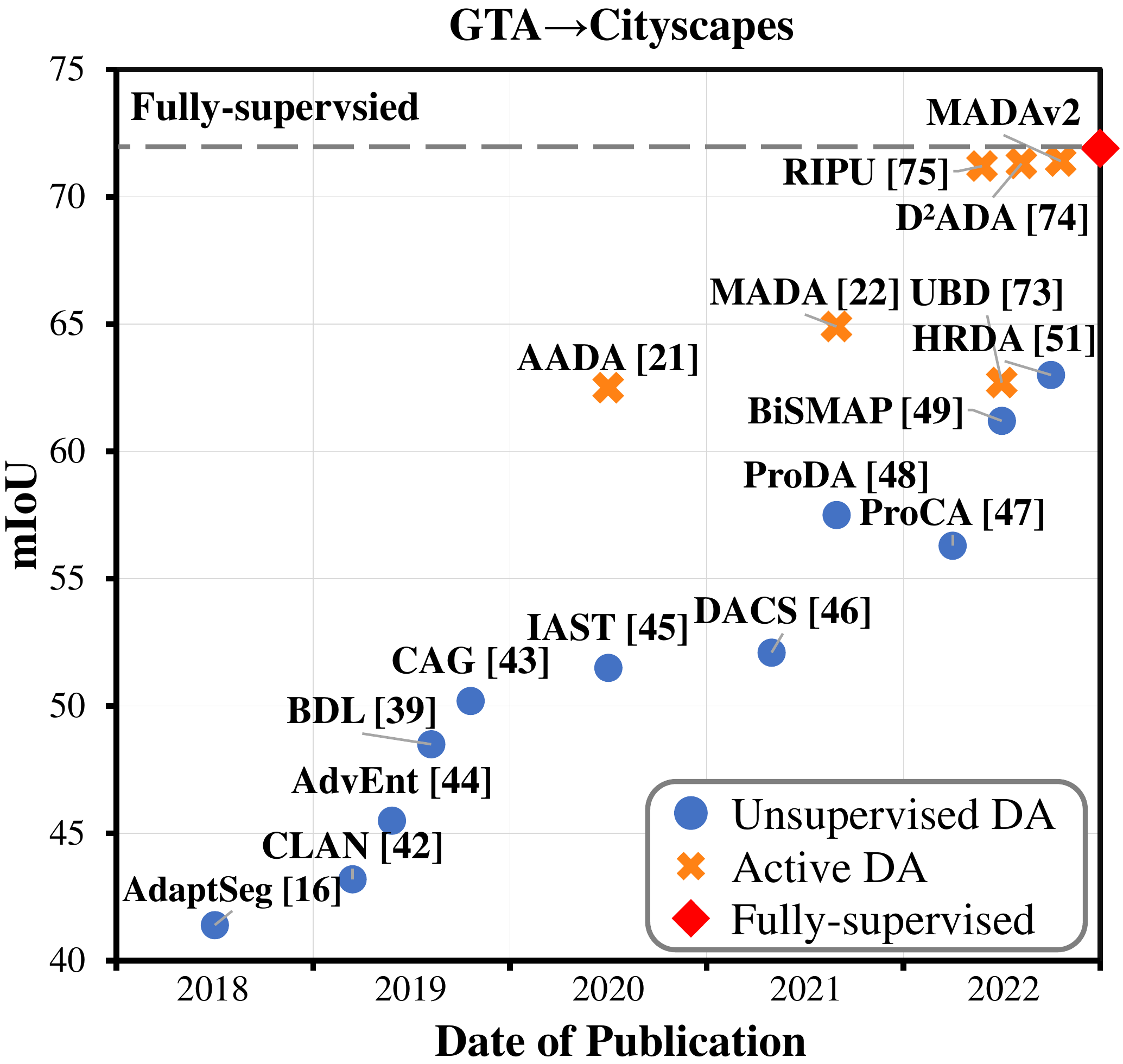}
	\caption{The mIoU of state-of-the-art DA methods on Gta5 $\rightarrow$ Cityscapes adaptation. All the results of active DA methods are trained additionally with 5\% annotation from the target domain. Our MADAv2 outperforms previous DA methods, and is very close to the fully-supervised upperbound.}
	\label{fig:miou}
\end{figure}

We visualize the generated target features with our MADAv2 using t-SNE in Fig.~\ref{fig:cluster_visual_1} as well as corresponding segmentation. It can be seen that MADAv2 effectively relieves the aforementioned distribution distortion problem and yields better segmentation. 
Also, as depicted in Fig.~\ref{fig:miou}, it outperforms previous domain adaptation works.
It achieves 71.4\% and 71.8\% mIoU on the segmentation of GTA5 and SYNTHIA datasets, respectively, improving upon the previous MADA method by 6.5\% and 3.7\% mIoU, and is close to the upperbound of 71.9\% and 73.0\%, on the GTA5 and SYNTHIA dataset respectively.
In summary, this study makes the following contributions:

\begin{itemize}

\item To the best of our knowledge, our work is the first study to adopt active learning to assist domain adaptation regarding the semantic segmentation tasks. By annotating a few target-domain samples, the distortion of the target-domain feature distribution can be effectively alleviated and superior segmentation performance can be achieved.

\item {We propose a new \textit{Dual\_Domain\_Distance} active metric, considering both the source and target domain distributions for selecting better active samples.
The source and target anchors are obtained to characterize the multimodal distributions of both domains. Then we select samples far from source and target anchors simultaneously, which are the most complementary to the source domain and informative in the target domain.}

\item 
We propose to effectively tackle the long-tail distribution problem by introducing the OHEM loss along with the adaptive cutmix and copy-paste augmentation in the new MADAv2 method, and learn better latent representation through the multi-anchor soft-alignment loss.


\end{itemize}

\section{Related Work}

\subsection{Unsupervised Domain Adaptation}
Unsupervised domain adaptation (UDA) is aimed at addressing the domain shift problem in a wide variety of computer vision tasks including classification~\cite{glorot2011domain, li2019joint, li2021transferable, long2018conditional}, detection~\cite{chen2018domain, vs2021mega}, and segmentation~\cite{tsai2018learning, liu2020open}.

Recent UDA methods can be roughly divided into two groups: maximum mean discrepancy (MMD) based~\cite{long2015learning,long2017deep,zellinger2017central,sun2016deep} and adversarial learning based~\cite{tsai2018learning,zhu2017unpaired,li2019bidirectional,chang2019all}.
The MMD kernel was first introduced in~\cite{long2015learning}, which measures the discrepancy of features from different domains quantitatively.
Subsequent studies proposed several improved MMD kernels for more accurate measurement of the domain discrepancy, including MK-MMD~\cite{long2015learning}, JMMD~\cite{long2017deep}, CMD~\cite{zellinger2017central} and CORAL~\cite{sun2016deep}. 
These kernels minimize the discrepancy to force features from different domains to align with each other, thus addressing the domain shift problem.
However, it is impractical to directly adopt the MMD-based methods in segmentation tasks, because these methods require complex computation in the high-dimension feature space.

In contrast, adversarial learning based methods are preferred for UDA of segmentation tasks, where the two domain distributions are drawn together via a domain discriminator. Among these methods, the classical appearance matching method CycleGAN~\cite{zhu2017unpaired} constructs two adversarial subnets to translate unpaired source and target images;
BDL~\cite{li2019bidirectional} leverages label consistency to improve the UDA performance;
DISE~\cite{chang2019all} proposes a disentangled representation learning architecture~\cite{huang2018multimodal} to preserve structural information during image translation.
In addition, feature aligning methods such as CLAN~\cite{luo2019taking} and CAG~\cite{zhang2019category} utilize category-based distribution alignment to adapt the source and target domains in the feature and output spaces. Another work
AdvEnt~\cite{vu2019advent} designs a novel loss function to maximize the prediction certainty in the target domain to boost the UDA performance.

To address the instability of adversarial UDA methods, IAST~\cite{mei2020instance} and DACS~\cite{tranheden2021dacs} apply a self-training (ST) framework to replace the adversarial training phase by learning from the refined pseudo labels. Based on the ST framework, ProCA~\cite{jiang2022prototypical}, ProDA~\cite{zhang2021prototypical} and BiSMAP~\cite{lu2022bidirectional} introduce the representative prototypes to provide additional restrictions.
Besides the training strategy, the Transformer~\cite{liu2021swin} is also utilized to improve the UDA semantic segmentation performance. With its powerful representation learning ability, UDA methods~\cite{hoyer2022hrda,hoyer2022daformer} based on a Transformer backbone show significantly better performance compared to those based on a ResNet backbone.

Despite the encouraging progress, UDA methods unconditionally force the distributions of the two domains to be similar, which may distort the underlying latent distribution of the target domain if it presents intrinsic difference from that of the source domain.
In this work, we address such distortion with active learning (AL) \cite{settles2009active}, at the price of only a little annotation workload.

\begin{figure*}[!htp]
	\centering
	\includegraphics[width=2.0\columnwidth]{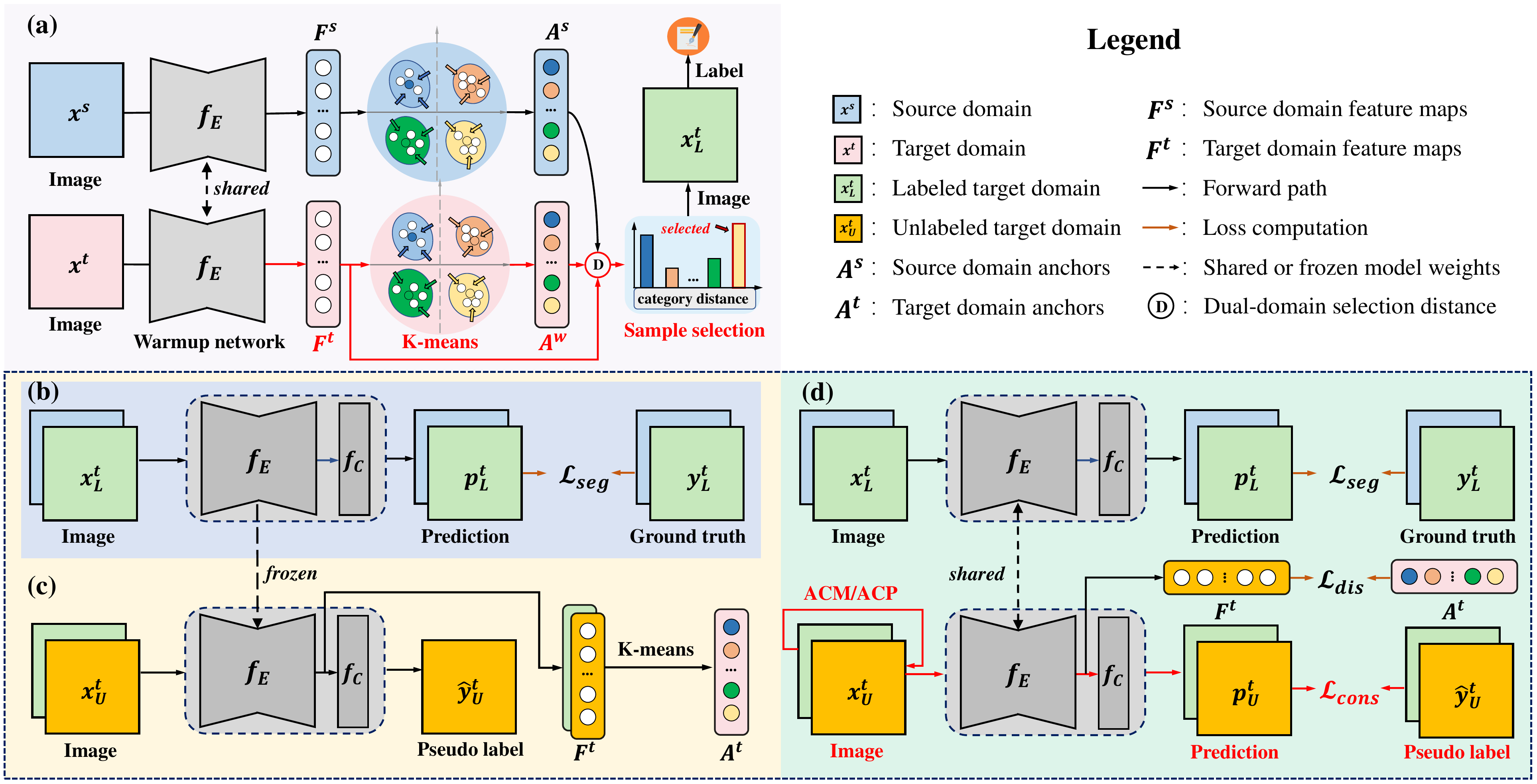}
	\caption{Overview of the proposed MADAv2 framework. The newly proposed components are highlight in red.}
	\label{fig:framework}
\end{figure*}

\subsection{Active Learning and Domain Adaptation}
AL aims at reaching the optimal performance at a low annotation cost by actively selecting a few samples that are most helpful to performance improvement if labeled~\cite{cohn1996active}.
Over the past decade, several sample selection strategies have been proposed for AL,
including uncertainty-based~\cite{lewis1994heterogeneous,scheffer2001active}, diversity-based~\cite{dutt2016active,hoi2009semisupervised}, representativeness-based~\cite{huang2010active,dasgupta2008hierarchical,nguyen2004active}, and expected model change based~\cite{freytag2014selecting,kading2015active,vezhnevets2012weakly} strategies. 
These strategies have been successfully applied to various computer vision tasks, such as image classification~\cite{qi2008two}, object detection~\cite{kao2018localization,Ji_2020_CoNet,Zhang_2020_LFNet}, and image segmentation~\cite{sun2015active}.
In this work, we argue that it is beneficial to introduce AL to the DA problem, to mitigate the distortion of the target-domain distribution.
AL entails only a little annotation cost, which is acceptable in many scenarios considering the potential performance gain.
Furthermore, with a proper sample selection strategy, AL can identify the samples most representative of the exclusive components in the target-domain distribution for annotation.
Hence, how to select the AL samples becomes a critical issue.

As far as the authors are aware of, only a few studies have attempted to apply AL to DA problems.
An early work by Chattopadhyay \emph{et al.}~\cite{chattopadhyay2013joint} 
uses the MMD distance between the source and target domains for active sample selection during the DA process.
However, it is practically prohibitive to apply MMD distances for segmentation DA problems, as mentioned earlier.
More recently, Huang \emph{et al.}~\cite{huang2018cost} proposed to fine-tune pre-trained models for classification tasks and involved additional active sample selection in every iteration.
In comparison, our framework takes a step forward to make dense predictions for segmentation tasks, and simplifies the active learning process to a one-time sample selection.
Being closely related to our work, Active Adversarial Domain Adaptation (AADA)~\cite{su2020active} proposed AL for DA with the adversarial learning~\cite{ganin2016domain} strategy, which selects representative samples by jointly considering diversity and uncertainty criteria.

In addition to the sample-wise active learning, pixel-wise~\cite{shin2021labor, you2022pixel, wu2022d} and region-wise~\cite{xie2022towards} active DA methods have also been proposed for semantic segmentation tasks. Shin \emph{et al.}~\cite{shin2021labor}, You \emph{et al.}~\cite{you2022pixel} and Xie \emph{et al.}~\cite{xie2022towards} selected pixels or regions by measuring prediction uncertainty. Wu \emph{et al.}~\cite{wu2022d} proposed a density-aware strategy to choose samples that
were representative in the target domain yet scarce the source domain. However, the annotators have to go through every image for annotating a few image areas, which costs much more time than only labeling a few images. Especially in some applications like medical images, the annotators need to analyze the whole image to determine the label of a specific region.

In this work, we propose to capture more comprehensive information from not only the target domain, but also the source domain. By modeling both the source and target distributions as multimodal (in contrast to the implicit unimodal assumption in previous works such as AADA), our method can achieve substantial performance improvement, as experimentally validated in Section~\ref{sec:sample_selection}.

\section{Proposed Approach}
In this section we elaborate on the proposed MADAv2 framework. An illustration of its overall structure is given in Fig. \ref{fig:framework}, with the newly proposed components being highlighted in red.  MADAv2 consists of two main stages: active target sample selection based on multiple anchors of both source and target domains (Fig. \ref{fig:framework}(a)) in Section~\ref{sec:multi-anchor}, and semi-supervised self-training aiming at exploiting information from both labeled and unlabeled data (Figs. \ref{fig:framework}(b), \ref{fig:framework}(c) and \ref{fig:framework}(d)) in Section~\ref{sec:semi-supervised-DA}.
Below we first formally define our problem setting and then explain the two stages in detail.

\subsection{Problem Setting}
The goal of semantic segmentation is to train a model $\mathbf{M}$ to map a sample $x$ in the image space $X$ to a prediction $y$ in the label space $Y$, where $x \in \mathbb{R}^{H \times W \times 3}$ with $H$ denoting the height, $W$ the width, and 3 the color channels, and $y \in\{0,1\}^{H \times W \times C}$ with $C$ denoting the number of segmentation categories.
For DA, there are $N_{s}$ image-label pairs $X^s=\left\{\left(x^{s}, y^{s}\right)\right\}$ in the source domain, and $N_{t}$ unlabeled images $X^t=\left\{x^{t}\right\}$ in the target domain.
For AL, $N_{a}$ active samples are selected in the target domain for annotation, where $N_{a} \ll N_t$, so that the target-domain data consist of $N_{a}$ image-label pairs $X^t_L=\left\{\left(x^{t}_{L}, y^{t}_{L}\right)\right\}$ and $N_{t}-N_{a}$ unlabeled images $X^t_U=\left\{x^{t}_{U}\right\}$.
The target of this work is to optimize the segmentation performance of $\mathbf{M}$ in the target domain while keeping $N_a$ small.

\subsection{Multi-anchor based Active Sample Selection}
\label{sec:multi-anchor}
\noindent
\textbf{Multiple Anchoring Mechanism.}
We propose an efficient anchoring mechanism to model the domain distributions, and shrink the gap between network predictions and the anchors by forming compact clusters around the anchors.

Previous works CAG~\cite{zhang2019category} and ProDA~\cite{zhang2021prototypical} average all image-level features of the source domain to obtain a centroid representing the entire domain, 
which implicitly assume a unimodal distribution.
In practice, however, the distribution of a domain often comprises more than a single mode \cite{cui2020unified}.
Although different images may contain the same categories of objects (\emph{e.g.}, road, car, human, and vegetable), they can be classified into various scenes (\emph{e.g.}, highway, uptown, and suburb). 

We then adopt \textit{multiple anchors} instead of a single centroid to characterize the domain distribution. By concatenating the features of different categories into an image-level `connected' vector, we perform clustering on them to estimate scene-specific representative distributions with cluster centers, denoted as `anchors’. We then select the most complementary and informative samples based on both the source and the target anchors.

As a warm-up model, we employ the common adversarial training~\cite{tsai2018learning} strategy to narrow the gap between the source and target domains.
Then we freeze the feature encoder $f_E$ and calculate the feature map $F^s_{c}(x^s)$ of a source sample $x^s$ for a certain category $c$ by:
\begin{equation}
    F_{c}^{s}(x^s)=\frac{1}{\left|\Lambda_{c}^{s}\right|}
    y_{c}^{s}\otimes
    \left.f_{E}\left(x^{s}\right)\right|_{c},
\end{equation}
where $y_{c}^{s}$ denotes the label map for category $c$,
$\left.f_{E}\left(x^{s}\right)\right|_{c}$ is the networks' output for category $c$,
$\otimes$ denotes element-wise multiplication for extracting category-exclusive information,
and $\left|\Lambda_{c}^{s}\right|$ is the number of pixels belonging to the specific category.
The final feature vector $F^{s}(x^s)$ of the source image $x^s$ is obtained by first flattening the $F_{c}^{s}(x)$ of each category into a vector followed by concatenating the vectors of all categories into a long vector.
Then, we apply the K-means method~\cite{macqueen1967some} to feature vectors of all source images to group them into $K$ clusters, by minimizing the following error:
\begin{equation}
    \sum_{k=1}^{K} \sum_{x \in \mathcal{C}_k}\left\|F^{s}(x^s)-A^s_{k}\right\|_{2}^{2},
    \label{eq:kmeans}
\end{equation}
where $\left\|\cdot\right\|_{2}^{2}$ denotes the $L2$ distance and $A^s_k$ is the centroid of the cluster $\mathcal{C}_k$:
\begin{equation}
    A^{s}_{k}=\frac{1}{\left|\mathcal{C}_{k}\right|} \sum_{x \in \mathcal{C}_{k}} F^{s}(x^s),
    \label{eq:centroid}
\end{equation}
where $\left|\mathcal{C}_{k}\right|$ denotes the number of images belonging to $\mathcal{C}_k$.
The centroids $\{A^s_k\}$ are used as the source-domain anchors, against which the target images will be compared for active sample selection. Note that the cluster number $K$ is not the same as the number of segmentation categories $C$, and the impact of different values of $K$ is explored in Section~\ref{sec:impact_of_K}.

\begin{figure*}[ht]
	\centering
	\includegraphics[width=1.5\columnwidth]{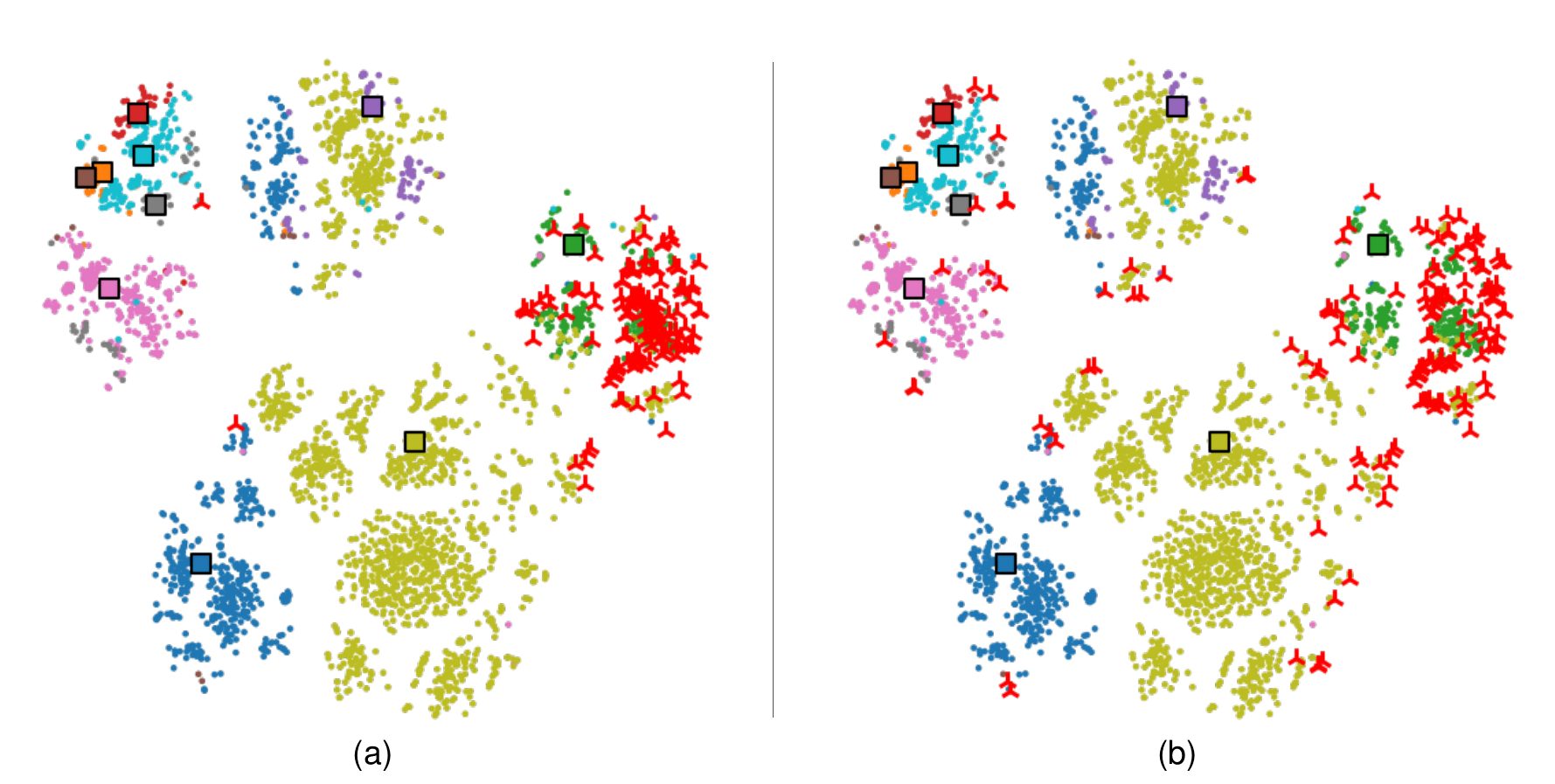}
	\caption{Visualization (t-SNE~\cite{van2008visualizing}) of different sample selection methods. In the scatters, dots of different colors denote the target samples of different clusters, squares represent corresponding cluster centroids, and \textit{red triangles} denote the selected AL samples. (a) The scatter of MADA~\cite{ning2021multi}, where we can find the active samples are crowded in the target domain. (b) The scatter of MADAv2, which considers the distance to both the source and target centers comprehensively.}
	\label{fig:scatters}
\end{figure*}

\noindent
\textbf{Active Target Sample Selection.}
For single-domain AL, uncertainty-based metrics have been extensively used to select the samples which are the most difficult to segment~\cite{siddiqui2020viewal}.
For multi-domain AL, however, we argue that the more dissimilar the target samples are to the source-domain, the more complementary they are to the segmentation network. 
Hence, we measure the dissimilarity by the distance between the target samples and the source anchors to assess the importance of unlabeled target samples for domain adaptation in our previous work~\cite{ning2021multi}. 
However, the selected samples may be \textit{crowded} in the target domain, as shown in Fig.~\ref{fig:scatters} (a). 
In addition, with a sufficient amount of data, the knowledge of the samples close to the target anchors can be learned through the semi-supervised learning strategy. 
To avoid information redundancy, we propose to additionally consider the distance of the target samples to the target anchors, so that the selected samples can provide as much information as possible.

Specifically, we first calculate the per category feature map of a target-domain image $x^t$:
\begin{equation}
    F_{c}^{t}(x^t)=\frac{1}{\left|\Lambda_{c}^{t}\right|}
    \hat{y}_{c}^{t}\otimes
    \left.f_{E}\left(x^{t}\right)\right|_{c},
    \label{eq:target_feature}
\end{equation}
where $\hat{y}_{c}^{t}$ is the predicted map for category $c$,
and $\left|\Lambda_{c}^{t}\right|$ is the number of pixels belonging to the specific category according to $\hat{y}_{c}^{t}$.
Then, we concatenate $F_{c}^{t}(x^t)$ of all categories to obtain the image-level feature vector $F^{t}(x^t)$.

Similar to the acquirement of source-domain anchors $\{A^s_k\}$ with Eq.~(\ref{eq:centroid}), we can obtain the target-domain anchors $\{A^w_k\}$ with the warm-up model and K-means clustering.
Eventually, we calculate the $L2$ distances from $F^{t}(x^t)$ to its nearest source anchor and target anchor, and define the sum of the two distances as the \textit{Dual\_Domain\_Distance}, with which the distances from sample $x^t$ to the source and target domains are considered comprehensively:
\begin{equation}
    D(x^t)=\min_k \left\|F^{t}(x^t)-A_{k}^{s}\right\|_{2}^{2} + \min_k \left\|F^{t}(x^t)-A_{k}^{w}\right\|_{2}^{2}.
\label{eq:distance}
\end{equation}
Here, the first item $\min_k \left\|F^{t}(x^t)-A_{k}^{s}\right\|_{2}^{2}$
aims to find the samples which are most complementary to the source domain, while the second item $\min_k \left\|F^{t}(x^t)-A_{k}^{w}\right\|_{2}^{2}$ favors the `\textit{outliers}' in the target domain. In this way, more information can be introduced for domain adaptation. 
Then, we select the samples with the largest \textit{Dual\_Domain\_Distance} as active samples and annotate them for subsequent training, hoping to learn unique characteristics of the target-domain distribution from these active annotations. As shown in Fig.~\ref{fig:scatters} (b), the samples selected in MADAv2 follow a better scatter in the target distribution.

\subsection{Semi-supervised Domain Adaptation}
\label{sec:semi-supervised-DA}
\noindent
\textbf{Step-1: Injecting Target-domain Specific Knowledge.}
The actively selected and annotated target-domain samples are added to the training process to learn information exclusive to the target domain (Fig. \ref{fig:framework}(b)).
Training data in this step consist of two parts: the labeled source samples $X^s$ and the active target samples $X^t_L$, and the model $f_E$ is fine-tuned with typical cross-entropy based segmentation losses:
\begin{equation}
    \mathcal{L}_{seg}=\mathcal{L}_{ce}\left(x^{s}, y^{s}\right)+\mathcal{L}_{ce}\left(x^{t}_{L}, y^{t}_{L}\right),
\label{eq:supervise_loss}
\end{equation}
where the cross-entropy loss $\mathcal{L}_{ce}$ is defined as:
\begin{equation}
    \mathcal{L}_{ce}=-\frac{1}{HW}\sum_{i=1}^{H \times W}\sum_{c=1}^{C}{y}_{i,c}\log\left(p_{i,c}\right),
\end{equation}
where $y_i$ denotes the label for pixel $i$ and $p_i$ is the probability predicted by the classifier $f_C$. 
As experimentally validated in Section~\ref{sec:sample_selection}, our multi-anchor based active sample selecting strategy is superior to previous strategies, and the model gets a steady improvement in performance with the actively selected samples.

\noindent
\textbf{Step-2: Constructing Target-domain Anchors and Pseudo Labels.}
To fully utilize the unlabeled target data $X^t_U$, we use the fine-tuned $f_{E}$ in step-1 to compute pseudo labels $\{\hat{y}^t_U\}$ for unlabeled target-domain samples as well as target-domain anchors $\{A^t_v\}_{v=1}^V$ (Fig. \ref{fig:framework}(c)), where $V$ represents the number of target-domain anchors.
Notably, as the target-domain anchors are a potentially biased estimation of the actual target-domain distribution, we naturally consider correcting them dynamically. 
As indicated by Xie \emph{et al.}~\cite{xie2016unsupervised}, re-clustering at each epoch may lead to the collapse of the training process due to jumps in cluster centroids between epochs.
Therefore, we treat the target-domain anchors as a memory bank, and employ the exponential moving average (EMA)~\cite{tarvainen2017mean} to progressively update each anchor in a smooth manner:
\begin{equation}
    A_{v}^{t}=\alpha A_{v}^{t}+(1-\alpha) F^{t}(x^t),
\label{eq:EMA}
\end{equation}
where $\alpha$ is set to 0.999 following~\cite{tarvainen2017mean}, and $F^{t}(x^t)$ is utilized to update the closest anchor. 
With both $\{\hat{y}^t_U\}$ and $\{A^t_v\}$ computed, we proceed to the next step for semi-supervised domain adaptation.

\begin{algorithm}[t!]
\caption{Advanced Multi-Anchor Based Active Domain Adaptation Segmentation (MADAv2)}
\label{alg:DCCS}
\begin{algorithmic}[1]\small
	    \renewcommand{\algorithmicrequire}{\textbf{Notation:}}
		\REQUIRE  Source-domain set $\left\{\left(x^{s}, y^{s}\right)\right\}$, selected active sample set $\left\{\left(x^{t}_{L}, y^{t}_{L}\right)\right\}$, unlabeled target-domain set $\left\{x^{t}_{U}\right\}$, encoder $f_E$, feature vector set of the source domain $\left\{F^{s}(x^s)\right\}$ and feature vector set of the target domain $\left\{F^{t}(x^t)\right\}$, and number of iterations $N$.\\
		\renewcommand{\algorithmicensure}{\textbf{Stage 1:}}
		\ENSURE
		\STATE Warm-up $f_E$ with adversarial training~\cite{tsai2018learning} to obtain $\left\{F^{s}(x^s)\right\}$ and $\left\{F^{t}(x^t)\right\}$.\\
		\STATE Apply K-means on $\left\{F^{s}(x^s)\right\}$  and $\left\{F^{t}(x^t)\right\}$ to group the source-domain features and warm-up target-domain features into $K$ clusters; \\
		\STATE Compute the centroids $\left\{A_k^s\right\}$ and  $\left\{A_k^w\right\}$  of the clusters  (Eq.~(\ref{eq:centroid})) to serve as the anchors of the source and target domains, respectively;  \\
		\STATE Calculate the distance from each target-domain sample to both $\left\{A_k^s\right\}$ and  $\left\{A_k^w\right\}$  (Eq.~(\ref{eq:distance})); \\
		\STATE Select 5\% target-domain samples with the smallest distances as active samples for annotation, resulting in set $\left\{\left(x^{t}_{L}, y^{t}_{L}\right)\right\}$. \\
		\renewcommand{\algorithmicensure}{\textbf{Stage 2:}}
		\ENSURE
		\STATE Fine-tune $f_E$ with both $\left\{\left(x^{s}, y^{s}\right)\right\}$ and $\left\{\left(x^{t}_{L}, y^{t}_{L}\right)\right\}$ by minimizing $\mathcal{L}_{seg}$  (Eq.~(\ref{eq:supervise_loss})), and re-obtain $\left\{F^{t}(x^t)\right\}$ with fine-tuned $f_E$;\\
		\STATE Initialize $\left\{A_v^t\right\}$ with K-means clustering on $\left\{F^{t}(x^t)\right\}$;\\
		\STATE \textbf{for} $i=1,...,N$ \textbf{do}\\
		\STATE \quad Calculate $\mathcal{L}_{seg}$ (Eq.~(\ref{eq:supervise_loss})) with $\left\{\left(x^{s}, y^{s}\right)\right\}$ and  $\left\{\left(x^{t}_{L}, y^{t}_{L}\right)\right\}$;\\
		\STATE \quad Calculate $\mathcal{L}_{cons}$ (Eq.~(\ref{eq:cons_loss})) and $\mathcal{L}_{dis}^t$ (Eq.~(\ref{eq:dis_loss})) with $\left\{x^{t}\right\}$;\\
		\STATE \quad Update $f_E$ by gradient descending $\nabla(\mathcal{L}_{seg}+\mathcal{L}_{cons}+\mathcal{L}_{dis}^t)$ (Eq.~(\ref{eq:overall_loss}));\\
		\STATE \quad Update $A_v^t$ with EMA (Eq.~(\ref{eq:EMA}));\\
		\STATE \textbf{end for}
	\end{algorithmic}
\end{algorithm}

\noindent
\textbf{Step-3: Semi-supervised Adaptation.}
\label{sec:semi}
Lastly, we combine the source data $X^s$, labeled target samples $X^t_L$, and unlabeled target samples $X^t_U$ for a semi-supervised training (\emph{i.e.}, a further fine-tuning of $f_E$) for domain adaptation (Fig. \ref{fig:framework}(d)).

In MADA \cite{ning2021multi}, we adopt a compact and efficient semi-supervised phase, but it cannot fully exploit the knowledge from unlabeled samples.
Due to the limited number of selected samples, the long-tail problem is much severer than the case in supervised semantic segmentation because of the small number of annotated pixels from tail classes.
Inspired by ST based semi-supervised methods~\cite{hu2021semi,wang2022semi}, we build a more powerful semi-supervised domain adaptation framework to alleviate the problem by focusing more on the difficult regions and pixels.

First, we introduce additional data augmentations, adaptive cutmix (ACM) and adaptive copy-paste (ACP)~\cite{hu2021semi}, to achieve region- and pixel-level augmentations. 
To be specific, given an unlabeled target sample $X^t_U$ and the corresponding pseudo label $\hat{y}_{U}^{t}$, the ACM selects less confidence pair $\left\{x_{U}^{t^{\prime}}, \hat{y}_{U}^{t^{\prime}}\right\}$ to replace some random regions of $\{x^t_U, \hat{y}_{U}^{t}\}$. It can be formulated as:
\begin{equation}
    \left\{\widehat{x_{U}^{t}}, \widehat{\hat{y}_{U}^{t}}\right\}=\operatorname{Cutmix}\left(\operatorname{Crop}\left(\left\{x_{U}^{t^{\prime}}, \hat{y}_{U}^{t^{\prime}}\right\}\right),\left\{x_{U}^{t}, \hat{y}_{U}^{t}\right\}\right),
\end{equation}
where $\{x^t_U, \hat{y}_{U}^{t}\}$ denotes a randomly chosen unlabeled target image and its corresponding pseudo label, $\left\{x_{U}^{t^{\prime}}, \hat{y}_{U}^{t^{\prime}}\right\}$ is elaborately chosen to improve the distribution of classes based on the confidence,
and $\left\{\widehat{x_{U}^{t}}, \widehat{\hat{y}_{U}^{t}}\right\}$ is generated by cropping a random region from $\left\{x_{U}^{t^{\prime}}, \hat{y}_{U}^{t^{\prime}}\right\}$ and covering the corresponding area in $\left\{x_{U}^{t}, \hat{y}_{U}^{t}\right\}$. 
Specifically, we first calculate the confidence of unlabeled samples as:
\begin{equation}
    \operatorname{Conf}_{c}=\frac{1}{\left|\Lambda_{c}^{t}\right|} \sum_{i=1}^{H \times W} \hat{y}_{i,c}^{t}\log\left(p_{i,c}\right) , \quad c \in\{1, \ldots, C\},
\end{equation}
where $C$ is the number of categories, $\left|\Lambda_{c}^{t}\right|$ denotes the number of pixels belonging to category $c$ according
to its pseudo label $\hat{y}^t_i$, and $log\left(p_{i,c}\right)$ denotes the $c$-th channel prediction of the $i$-th pixel.

With the obtained confidence, the probability of sample $x_{U}^{t^{\prime}}$ to be chosen to provide crops for ${x^t_U}$ is defined as:
\begin{equation}
    \boldsymbol{r}=\sum_{c=1}^{C}\operatorname{Softmax}(1-\operatorname{Conf}_{c})
    \label{eq:prob}.
\end{equation}

For labeled active samples $X^t_L$, we utilize the adaptive copy-paste method to copy the pixels of non-long-tail classes to long-tail classes (\emph{e.g.}, copy cars to the road, and persons to the sidewalk), which can be formulated as:
\begin{equation}
    \{\widehat{x^t_L}, \widehat{y^t_L}\}=\operatorname{Paste}\left(\operatorname{Copy}\left(\{{x^{t^{\prime}}_L}, {y^{t^{\prime}}_L}\}\right), \{x^t_L, y^t_L\}\right),
\end{equation}
where $\{{x^{t^{\prime}}_L}, {y^{t^{\prime}}_L}\}$ and $\{x^t_L, y^t_L\}$ denote active image-label pairs, $\{\widehat{x^t_L}, \widehat{y^t_L}\}$ is the augmented image.
The procedure is similar to ACM, except that precise pixels instead of random regions are selected to replace the ones in the corresponding pairs.
Based on the augmented training pairs, the consistency loss can be defined as:
\begin{equation}
    \mathcal{L}_{cons} = \mathcal{L}_{ohem}(\widehat{x^t_L}, \widehat{y^t_L}) + \mathcal{L}_{ohem}(\widehat{x^t_U}, \widehat{\hat{y}^t_U}).
    \label{eq:cons_loss}
\end{equation}
The online hard example mining (OHEM) loss is a masked cross-entropy loss, where only the pixels with low probabilities are considered. For more details regarding the pixel selection, please refer to the original study~\cite{shrivastava2016training}.

Beyond the self-training period, we propose a novel soft alignment loss to explicitly shrink the gap between the sample features and the anchors in the target domain:

\newcommand{\slfrac}[2]{\left.#1\middle/#2\right.} 
\begin{equation}
    \mathcal{L}_{dis}^{t}= {V}\Big/{{\sum}_{v=1}^{V}\frac{1}{\left\|F^{t}(x^t)-A_{v}^{t}\right\|_{2}^{2}}}.
\label{eq:dis_loss}
\end{equation}
Intuitively, by minimizing the soft alignment loss, features of the target-domain samples output by the model are drawn towards the target-domain anchors, encouraging a more faithful learning of the underlying target-domain distribution represented by these anchors.

Thus, the overall loss function for the semi-supervised learning can be formulated as:
\begin{equation}
        \mathcal{L}_{semi}=\mathcal{L}_{seg}+\mathcal{L}_{cons}+\mathcal{L}_{dis}^{t}.
\label{eq:overall_loss}
\end{equation}
The entire training pipeline is summarized in Algorithm \ref{alg:DCCS}.

\begin{table*}
    \caption{Comparison with other DA methods on (a) GTA5 to Cityscapes and (b) SYNTHIA to Cityscapes adaptation task. Best results are shown in \textbf{bold}. The 'mIoU$^{\bigstar}$' denotes the average mIoU of 13 classes in~\cite{tsai2018learning}. The 'MADAv2*' means re-implementing our method with the setting of D$^2$ADA~\cite{wu2022d} for a fair comparison.}
    \label{table:comparison-GTA5}
    \centering
    \setlength\tabcolsep{1.9mm}
    \renewcommand\arraystretch{1.1}
	\def \mysp {\hspace{7pt}}
    {\small 
    \scalebox{0.9}{
    \begin{tabular}{l@{\ }@{\ }b{0.35cm}b{0.35cm}b{0.35cm}b{0.35cm}b{0.35cm}b{0.35cm}b{0.35cm}b{0.35cm}b{0.35cm}b{0.35cm}b{0.35cm}b{0.35cm}b{0.35cm}b{0.35cm}b{0.35cm}b{0.35cm}b{0.35cm}b{0.35cm}b{0.35cm}c}
    \toprule[1pt]
    \multicolumn{21}{c}{(a) GTA5 $\rightarrow$ Cityscapses} \\
    \toprule
        Method &\rotatebox{45}{road} &\rotatebox{45}{sidewalk} &\rotatebox{45}{building} &\rotatebox{45}{wall} &\rotatebox{45}{fence} &\rotatebox{45}{pole} &\rotatebox{45}{light} &\rotatebox{45}{sign} &\rotatebox{45}{veg} &\rotatebox{45}{terrain} &\rotatebox{45}{sky} &\rotatebox{45}{person} &\rotatebox{45}{rider} &\rotatebox{45}{car} &\rotatebox{45}{truck} &\rotatebox{45}{bus} &\rotatebox{45}{train} &\rotatebox{45}{mbike} &\rotatebox{45}{bicycle} &mIoU  \\
    \toprule
        AdaptSeg~\cite{tsai2018learning} &86.5 &25.9 &79.8 &22.1 &20.0 &23.6 &33.1 &21.8 &81.8 &25.9 &75.9 &57.3 &26.2 &76.3 &29.8 &32.1 &7.2 &29.5 &32.5 &41.4 \\
        CLAN~\cite{luo2019taking} &87.0 &27.1 &79.6 &27.3 &23.3 &28.3 &35.5 &24.2 &83.6 &27.4 &74.2 &58.6 &28.0 &76.2 &33.1 &36.7 &6.7 &31.9 &31.4 &43.2 \\
        AdvEnt~\cite{vu2019advent} &89.4 &33.1 &81.0 &26.6 &26.8 &27.2 &33.5 &24.7 &83.9 &36.7 &78.8 &58.7 &30.5 &84.8 &38.5 &44.5 &1.7 &31.6 &32.4 &45.5 \\
        BDL~\cite{li2019bidirectional} &91.0 &44.7 &84.2 &34.6 &27.6 &30.2 &36.0 &36.0 &85.0 &43.6 &83.0 &58.6 &31.6 &83.3 &35.3 &49.7 &3.3 &28.8 &35.6 &48.5 \\
        CAG~\cite{zhang2019category} &90.4 &51.6 &83.8 &34.2 &27.8 &38.4 &25.3 &48.4 &85.4 &38.2 &78.1 &58.6 &34.6 &84.7 &21.9 &42.7 &41.1 &29.3 &37.2 &50.2 \\
        IAST~\cite{mei2020instance} &93.8 &57.8 &85.1 &39.5 &26.7 &26.2 &43.1 &34.7 &84.9 &32.9 &88.0 &62.6 &29.0 &87.3 &39.2 &49.6 &23.2 &34.7 &39.6 &51.5 \\
        DACS~\cite{tranheden2021dacs} &89.9 &39.7 &87.9 &30.7 &39.5 &38.5 &46.4 &52.8 &88.0 &44.0 &88.8 &67.2 &35.8 &84.5 &45.7 &50.2 &0.0 &27.3 &34.0 &52.1 \\
        ProCA~\cite{jiang2022prototypical} &91.9 &48.4 &87.3 &41.5 &31.8 &41.9 &47.9 &36.7 &86.5 &42.3 &84.7 &68.4 &43.1 &88.1 &39.6 &48.8 &40.6 &43.6 &56.9 &56.3 \\
        ProDA~\cite{zhang2021prototypical} &87.8 &56.0 &79.7 &46.3 &44.8 &45.6 &53.5 &53.5 &88.6 &45.2 &82.1 &70.7 &39.2 &88.8 &45.5 &59.4 &1.0 &48.9 &56.4 &57.5 \\
        BiSMAP~\cite{lu2022bidirectional} &89.2 &54.9 &84.4 &44.1 &39.3 &41.6 &53.9 &53.5 &88.4 &45.1 &82.3 &69.4 &41.8 &90.4 &56.4 &68.8 &51.2 &47.8 &60.4 &61.2 \\
        HRDA~\cite{hoyer2022hrda}  &96.2 &73.1 &89.7 &43.2 &39.9 &47.5 &60.0 &60.0 &89.9 &47.1 &90.2 &75.9 &49.0 &91.8 &61.9 &59.3 &10.2 &47.0 &65.3 &63.0 \\
    \toprule
    \toprule
        AADA (5\%)~\cite{su2020active} &94.1 &66.7 &87.7 &43.0 &49.9 &49.4 &54.8 &59.8 &89.3 &47.8 &89.7 &72.5 &43.0 &90.7 &51.8 &48.4 &41.8 &41.5 &66.3 &62.5 \\
        UBD (5\%)~\cite{you2022pixel} &95.4 &70.8 &88.4 &44.9 &48.3 &45.2 &49.5 &58.2 &88.3 &44.6 &91.0 &72.6 &37.3 &90.7 &58.5 &65.7 &44.2 &33.7 &64.7 &62.7 \\
        MADA (5\%)~\cite{ning2021multi} &95.1 &69.8 &88.5 &43.3 &48.7 &45.7 &53.3 &59.2 &89.1 &46.7 &91.5 &73.9 &50.1 &91.2 &60.6 &56.9 &48.4 &51.6 &68.7 &64.9 \\
        RIPU (5\%)~\cite{xie2022towards} &97.0 &77.3 &\textbf{90.4} &\textbf{54.6} &\textbf{53.2} &47.7 &55.9 &64.1 &90.2 &\textbf{59.2} &93.2 &75.0 &54.8 &92.7 &73.0 &79.7 &\textbf{68.9} &55.5 &70.3 &71.2 \\
        MADAv2 (5\%) &\textbf{97.4} &\textbf{79.8} &90.3 &46.6 &52.5 &\textbf{54.3} &\textbf{62.3} &\textbf{71.7} &\textbf{91.2} &50.2 &\textbf{94.0} &\textbf{77.8} &\textbf{56.8} &\textbf{93.2} &\textbf{74.7} &\textbf{80.3} &52.4 &\textbf{57.5} &\textbf{73.6} &\textbf{71.4} \\
        \toprule
        D$^2$ADA* (5\%)~\cite{wu2022d} &97.0 &77.8 &90.0 &\textbf{46.0} &\textbf{55.0} &52.7 &58.7 &65.8 &90.4 &\textbf{58.9} &92.1 &75.7 &54.4 &92.3 &69.0 &78.0 &\textbf{68.5} &\textbf{59.1} &72.3 &71.3 \\
        MADAv2* (5\%) &\textbf{97.2} &\textbf{78.4} &\textbf{90.5} &45.2 &53.9 &\textbf{55.0} &\textbf{61.8} &\textbf{72.2} &\textbf{91.2} &51.1 &\textbf{93.9} &\textbf{78.3} &\textbf{57.4} &\textbf{93.5} &\textbf{76.6} &\textbf{79.6} &53.6 &58.0 &\textbf{73.8} &\textbf{71.6} \\
        \toprule
        Fully-supervised &97.7 &82.0 &90.7 &55.2 &56.6 &53.1 &58.6 &67.5 &91.3 &60.4 &93.8 &75.7 &52.7 &93.1 &76.1 &78.5 &58.4 &55.4 &69.5 &71.9 \\
    \bottomrule[1pt]
    \end{tabular}}
    \vspace{0.3cm}
    }
    
    {\small 
    \scalebox{0.9}{
    \begin{tabular}{l@{\ }@{\ }b{0.4cm}b{0.4cm}b{0.4cm}b{0.4cm}b{0.4cm}b{0.4cm}b{0.4cm}b{0.4cm}b{0.4cm}b{0.4cm}b{0.4cm}b{0.4cm}b{0.4cm}b{0.4cm}b{0.4cm}b{0.4cm}cc}
    \toprule[1pt]
    \multicolumn{19}{c}{(b) SYNTHIA $\rightarrow$ Cityscapses} \\
    \toprule
        Method &\rotatebox{45}{road} &\rotatebox{45}{sidewalk} &\rotatebox{45}{building} &\rotatebox{45}{wall} &\rotatebox{45}{fence} &\rotatebox{45}{pole} &\rotatebox{45}{light} &\rotatebox{45}{sign} &\rotatebox{45}{veg}  &\rotatebox{45}{sky} &\rotatebox{45}{person} &\rotatebox{45}{rider} &\rotatebox{45}{car}  &\rotatebox{45}{bus}  &\rotatebox{45}{mbike} &\rotatebox{45}{bicycle} &mIoU &mIoU$^{\bigstar}$  \\
    \toprule
        AdaptSeg~\cite{tsai2018learning} &79.2 &37.2 &78.8 &- &- &- &9.9 &10.5 &78.2 &80.5 &53.5 &19.6 &67.0 &29.5 &21.6 &31.3 &- &45.9 \\
        CLAN~\cite{luo2019taking} &81.3 &37.0 &80.1 &- &- &- &16.1 &13.7 &78.2 &81.5 &53.4 &21.2 &73.0 &32.9 &22.6 &30.7 &- &47.8 \\
        AdvEnt~\cite{vu2019advent} &85.6 &42.2 &79.7 &8.7 &0.4 &25.9 &5.4 &8.1 &80.4 &84.1 &57.9 &23.8 &73.3 &36.4 &14.2 &33.0 &41.2 &48.0 \\
        BDL~\cite{li2019bidirectional} &86.0 &46.7 &80.3 &- &- &- &14.1 &11.6 &79.2 &81.3 &54.1 &27.9 &73.7 &42.2 &25.7 &45.3 &- &51.4 \\
        CAG~\cite{zhang2019category} &84.7 &40.8 &81.7 &7.8 &0.0 &35.1 &13.3 &22.7 &84.5 &77.6 &64.2 &27.8 &80.9 &19.7 &22.7 &48.3 &44.5 &50.9 \\
        IAST~\cite{mei2020instance} &81.9 &41.5 &83.3 &17.7 &4.6 &32.3 &30.9 &28.8 &83.4 &85.0 &65.5 &30.8 &86.5 &38.2 &33.1 &52.7 &49.8 &57.0 \\
        DACS~\cite{tranheden2021dacs} &80.6 &25.1 &81.9 &21.5 &2.9 &37.2 &22.7 &24.0 &83.7 &90.8 &67.6 &38.3 &82.9 &38.9 &28.5 &47.6 &48.3 &54.8 \\
        ProCA~\cite{jiang2022prototypical} &90.5 &52.1 &84.6 &29.2 &3.3 &40.3 &37.4 &27.3 &86.4 &85.9 &69.8 &28.7 &88.7 &53.7 &14.8 &54.8 &53.0 &59.6 \\
        ProDA~\cite{zhang2021prototypical} &87.8 &45.7 &84.6 &37.1 &0.6 &44.0 &54.6 &37.0 &88.1 &84.4 &74.2 &24.3 &88.2 &51.1 &40.5 &45.6 &55.5 &62.0 \\
        BiSMAP~\cite{lu2022bidirectional} &81.9 &39.8 &84.2 &- &- &- &41.7 &46.1 &83.4 &88.7 &69.2 &39.3 &80.7 &51.0 &51.2 &58.8 &- &62.8 \\
        HRDA~\cite{hoyer2022hrda} &85.8 &47.3 &87.3 &27.3 &1.4 &50.5 &57.8 &61.0 &87.4 &89.1 &\textbf{76.2} &48.5 &87.3 &49.3 &55.0 &68.2 &61.2 &69.2 \\
    \toprule
    \toprule
        AADA (5\%)~\cite{su2020active} &93.9 &66.3 &87.5 &36.9 &41.1 &47.5 &53.1 &59.8 &88.6 &92.9 &71.3 &42.4 &90.0 &52.8 &34.8 &66.6 &64.1 &69.2 \\
        UBD (5\%)~\cite{you2022pixel}  &94.8 &70.8 &88.1 &39.4 &44.9 &43.7 &50.9 &57.9 &91.5 &88.8 &69.5 &50.2 &89.0 &67.2 &46.3 &68.4 &66.3 &71.8 \\
        MADA (5\%)~\cite{ning2021multi} &96.5 &74.6 &88.8 &45.9 &43.8 &46.7 &52.4 &60.5 &89.7 &92.2 &74.1 &51.2 &90.9 &60.3 &52.4 &69.4 &68.1 &73.3 \\
        RIPU (5\%)~\cite{xie2022towards} &97.0 &78.9 &\textbf{89.9} &47.2 &\textbf{50.7} &48.5 &55.2 &63.9 &\textbf{91.1} &93.0 &74.4 &\textbf{54.1} &\textbf{92.9} &\textbf{79.9} &55.3 &71.0 &71.4 &76.7 \\
        MADAv2 (5\%) &\textbf{97.3} &\textbf{80.4} &\textbf{89.9} &\textbf{49.2} &42.6 &\textbf{52.9} &\textbf{60.7} &\textbf{70.4} &90.8 &\textbf{94.1} &\textbf{75.2} &51.6 &92.3 &73.3 &\textbf{55.4} &\textbf{73.3} &\textbf{71.8} &\textbf{77.2} \\
    \toprule
        D$^2$ADA* (5\%)~\cite{wu2022d} &96.7 &76.8 &\textbf{90.3} &\textbf{48.7} &51.1 &54.2 &58.3 &68.0 &90.4 &93.4 &77.4 &\textbf{56.4} &92.5 &\textbf{77.5} &\textbf{58.9} &\textbf{73.3} &72.7 &77.7 \\
        MADAv2* (5\%) &\textbf{97.1} &\textbf{78.0} &\textbf{90.3} &42.8 &\textbf{52.0} &\textbf{54.6} &\textbf{62.7} &\textbf{70.4} &\textbf{91.2} &\textbf{93.7} &\textbf{78.0} &55.5 &\textbf{92.7} &76.2 &57.4 &73.0 
        &\textbf{72.8} &\textbf{78.1} \\
    \toprule
        Fully-supervised &97.7 &82.3 &90.8 &53.5 &57.3 &52.8 &58.9 &67.4 &91.4 &93.4 &75.7 &53.4 &92.6 &75.9 &55.1 &70.6 &73.0 &77.3 \\
    \bottomrule[1pt]
	\end{tabular}}
    }
\end{table*}

\section{Experiments}

\subsection{Datasets}
To demonstrate the superiority of our proposed method, two challenging \textit{synthetic-to-real} adaptation tasks, \emph{i.e.},  GTA5~\cite{richter2016playing} $\rightarrow$ Cityscapes~\cite{cordts2016cityscapes} and SYNTHIA~\cite{ros2016synthia} $\rightarrow$ Cityscapes are applied for evaluation. To be specific:

\begin{itemize}
\item GTA5 $\rightarrow$ Cityscapes: The GTA5 dataset consists of 24,966 synthetic images with 19-class segmentation, which is consistent with the Cityscapses dataset.
\item SYNTHIA $\rightarrow$ Cityscapes: Following the previous study~\cite{li2019bidirectional}, the SYNTHIA-RAND-CITYSCAPES set with 9,400 synthetic images containing 16-class segmentation is utilized for training.
\end{itemize}

In both settings, Cityscapes serves as the target domain, with 2,975 images for training and 500 images for evaluation. The segmentation performance is measured with the mean-Intersection-over-Union (mIoU)~\cite{everingham2015pascal} metric.

\subsection{Implementation Details}
We employ backbone DeepLab v3+~\cite{chen2018encoder} as the feature extractor $f_{E}$, which is composed of the ResNet-101~\cite{he2016deep} pretrained on ImageNet~\cite{deng2009imagenet} and the Atrous Spatial Pyramid Pooling (ASPP) module. The classifier $f_{C}$ is a typical convolutional layer with $C$ channels and 1 $\times$ 1 kernel size to transform the latent representation to semantic segmentation.
During the warm-up, the discriminator $f_{D}$ consists of 5 convolutional layers of kernel size 3 $\times$ 3 and stride 2 with numbers of filters set to $\{64, 128, 256, 512, 1\}$. 
The first three convolutional layers are followed with a Rectified Linear Unit (ReLU) layer, while the fourth one is followed by a leaky ReLU~\cite{maas2013rectifier} parameterized by 0.2. The proposed method is implemented on PyTorch with an NVIDIA Tesla V100 GPU. The input images are randomly resized with a ratio in $[0.5, 1.5]$ and then randomly cropped to 896 $\times$ 512 pixels. 

For warm-up, we train the model for 20 epochs in an adversarial manner with a cross entropy loss and an adversarial loss weighted by $0.01$. For the semi-supervised domain adaptation stage, we use the SGD optimizer to train our model for 200 epochs. The learning rate is initially set to $2.5 \times 10^{-4}$ and decayed by the poly learning rate policy with a power of 0.9.

Except for the comparison study in Section~\ref{sec:impact_of_sample_num}, we select 5\% target-domain samples as active samples for all experiments, which cost a little annotation workload but bring large performance gains.

\begin{figure*}[ht]
	\centering
	\includegraphics[width=1.9\columnwidth]{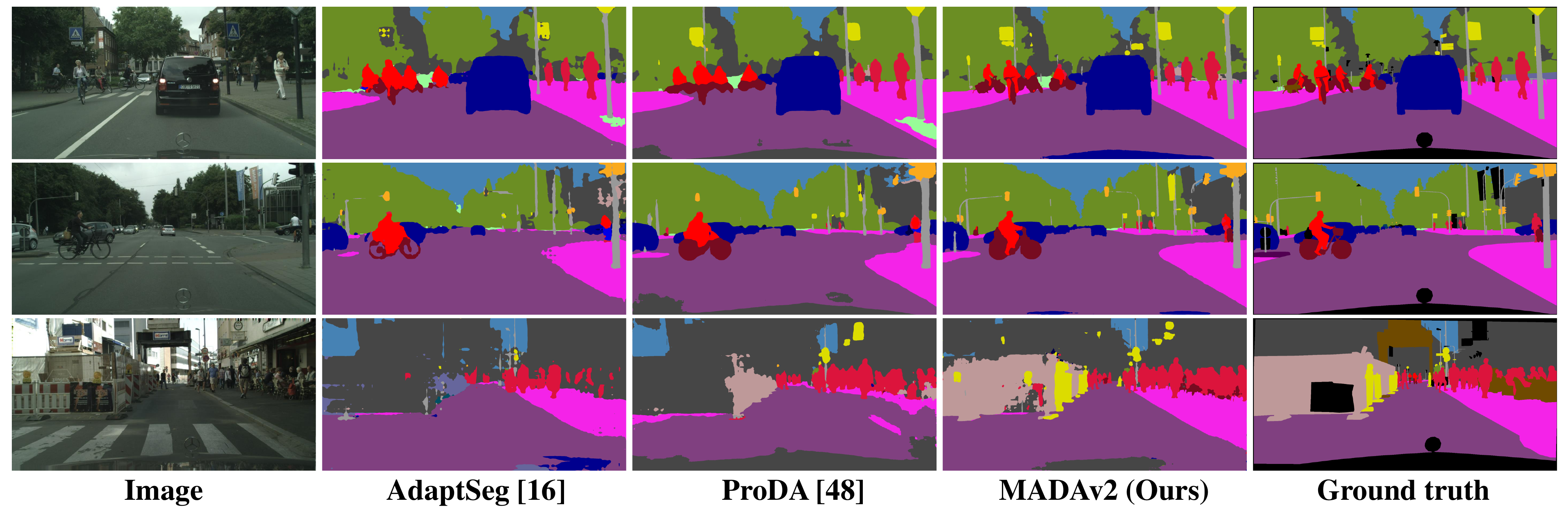}
	\caption{Qualitative results of DA segmentation for GTA5 $\rightarrow$ Cityscapes. For each image, we show the results of a typical adversarial method (i.e., AdaptSeg ~\cite{tsai2018learning}), typical unimodal prototype method (i.e., ProDA~\cite{zhang2021prototypical}) and MADAv2, respectively. The black region in 'Ground truth' is excluded from evaluation because it does not belong to any of the 19 classes.}
	\label{fig:visualizatoin}
\end{figure*}

\subsection{Main Results}
\label{sec:comparison}
As presented in Table~\ref{table:comparison-GTA5}, the proposed framework is compared with a series of unsupervised (UDA) and active DA methods.
For the UDA task, traditional adversarial-based~\cite{tsai2018learning,luo2019taking,vu2019advent,li2019bidirectional,zhang2019category}, prototype-ST-based~\cite{zhang2021prototypical,jiang2022prototypical,lu2022bidirectional} and the ResNet-101 version of HRDA~\cite{hoyer2022hrda} (which is the current SOTA) UDA methods are included for comparison (Transformer based methods are not listed due to the different feature extraction ability and segmentation performance upperbound). For the active DA task, we compare with the sample-based~\cite{su2020active}, pixel-based~\cite{you2022pixel,wu2022d} and region-based~\cite{xie2022towards} approaches using the same amount of annotation. The results of MADA~\cite{ning2021multi} are also listed.
As expected, we observe substantial improvements of our proposed method MADAv2 over the compared UDA methods, which indicates that with elaborately selected active samples, a little manual annotation workload can lead to large performance gains. 
In addition, MADAv2 outperforms another sample-based active DA method, \emph{i.e.}, AADA~\cite{su2020active}, by a large margin (8.9\% mIOU), demonstrating the effectiveness of the proposed multi-anchor strategy.
And the improvement compared to MADA demonstrates that the new sample selection metric and ST based semi-supervised domain adaptation can effectively address the weaknesses of the previous method. In addition, our method consistently shows better performance than the UBD~\cite{you2022pixel} and RIPU~\cite{xie2022towards}, which demonstrates that selecting a few images to annotate is better than labeling a few regions of each sample, despite the more time latter one costs. Note that MADAv2 achieves the best performance in most cases, except for the SYNTHIA to Cityscapes adaptation task with D$^2$ADA~\cite{wu2022d} due to different training settings; but with the same setting, MADAv2 shows better performance as indicated by '*' in Table~\ref{table:comparison-GTA5}.

The visualization of three example images is displayed in Fig.~\ref{fig:visualizatoin} for qualitative comparison. We can observe that by alleviating the distortion of target features, fewer segmentation errors as well as more precise boundaries can be obtained with the proposed MADAv2 method.

\begin{table}
    \caption{Ablation study. G $\rightarrow$ C denotes the GTA5 $\rightarrow$ Cityscapes and S $\rightarrow$ C denotes the SYNTHIA $\rightarrow$ Cityscapes adaptation.}
    \label{table:ablation_study}
    \centering
    \setlength\tabcolsep{1.9mm}
    \renewcommand\arraystretch{1.1}
	\def \mysp {\hspace{7pt}}
    {\small 
    \scalebox{1.0}{
    \begin{tabular}{b{1.4cm}b{0.5cm}b{0.5cm}b{0.5cm}b{0.5cm}b{0.5cm}cc}
    \toprule[1pt]
    & & & & & &G $\rightarrow$ C &S $\rightarrow$ C \\
    \toprule
         Method &A &B &C &D &E &mIoU &mIoU  \\
         \toprule
         $\mathbf{M}^{(0)}$ & & & & & &42.5 &42.9\\
         \toprule
         $\mathbf{M}^{(1)}$  &\checkmark & & & & &65.1 &65.2\\
         \toprule
         $\mathbf{M}^{(2)}$  &\checkmark &\checkmark & & & &68.5 &69.0\\
         $\mathbf{M}^{(3)}$  &\checkmark &\checkmark &\checkmark & & &69.5 &70.3\\
        $\mathbf{M}^{(4)}$  &\checkmark &\checkmark &\checkmark &\checkmark & &70.8 &71.1\\
        $\mathbf{M}^{(5)}$  &\checkmark &\checkmark &\checkmark &\checkmark &\checkmark &71.4 &71.8\\
    \toprule
    $\mathbf{M}^{(u)}$ & & & & & &71.9 &73.0\\
    \toprule
    \multicolumn{7}{l}{A: Training with active samples}\\
    \multicolumn{7}{l}{B: Prediction consistency loss}\\
    \multicolumn{7}{l}{C: OHEM loss}\\
    \multicolumn{7}{l}{D: ACM and ACP}\\
    \multicolumn{7}{l}{E: Soft-anchor alignment loss}\\
    \bottomrule[1pt]
    \end{tabular}}
    }
\end{table}

\begin{table*}[!htp]
    \caption{Experiments with different active sample selection methods. Best results are shown in \textbf{bold}.}
    \label{table:selection}
    \centering
    \setlength\tabcolsep{1.9mm}
    \renewcommand\arraystretch{1.1}
	\def \mysp {\hspace{7pt}}
    {\small 
    \scalebox{1.0}{
    \begin{tabular}{l@{\ }@{\ }b{0.35cm}b{0.35cm}b{0.35cm}b{0.35cm}b{0.35cm}b{0.35cm}b{0.35cm}b{0.35cm}b{0.35cm}b{0.35cm}b{0.35cm}b{0.35cm}b{0.35cm}b{0.35cm}b{0.35cm}b{0.35cm}b{0.35cm}b{0.35cm}b{0.35cm}c}
    \toprule[1pt]
    \multicolumn{21}{c}{GTA5 $\rightarrow$ Cityscapses} \\
    \toprule
        Method &\rotatebox{45}{road} &\rotatebox{45}{sidewalk} &\rotatebox{45}{building} &\rotatebox{45}{wall} &\rotatebox{45}{fence} &\rotatebox{45}{pole} &\rotatebox{45}{light} &\rotatebox{45}{sign} &\rotatebox{45}{veg} &\rotatebox{45}{terrain} &\rotatebox{45}{sky} &\rotatebox{45}{person} &\rotatebox{45}{rider} &\rotatebox{45}{car} &\rotatebox{45}{truck} &\rotatebox{45}{bus} &\rotatebox{45}{train} &\rotatebox{45}{mbike} &\rotatebox{45}{bicycle} &mIoU  \\
    \toprule
        Random  &94.8 &71.0 &86.6 &39.0 &43.5 &47.1 &55.2 &59.1 &88.7 &43.5 &87.1 &72.2 &35.8 &90.1 &51.1 &49.6 &29.8 &50.1 &66.2 &61.1\\
        Adversarial~\cite{tsai2018learning} &93.2 &65.3 &86.9 &38.5 &47.8 &48.3 &55.0 &56.2 &89.3 &46.5 &90.5 &71.9 &42.5 &90.6 &50.5 &47.4 &41.7 &41.0 &66.1 &61.5\\
        Entropy~\cite{vu2019advent} &94.7 &68.2 &87.8 &44.4 &47.5 &46.7 &49.9 &57.5 &89.4 &48.5 &90.9 &72.8 &43.4 &90.7 &54.6 &49.4 &30.3 &51.6 &66.4 &62.4\\
        AADA~\cite{su2020active} &94.1 &66.7 &87.7 &43.0 &49.9 &49.4 &54.8 &59.8 &89.3 &47.8 &89.7 &72.5 &43.0 &90.7 &51.8 &48.4 &41.8 &41.5 &66.3 &62.5\\
        Prototype~\cite{zhang2021prototypical} &95.6 &71.5 &87.9 &36.6 &47.6 &49.4 &53.5 &60.9 &88.5 &48.9 &91.0 &74.1 &48.4 &91.3 &51.7 &57.4 &22.5 &52.4 &68.9 &63.1\\
        MADA~\cite{ning2021multi} &93.5 &66.7 &86.8 &31.5 &46.3 &50.4 &57.5 &62.5 &88.8 &43.4 &91.2 &75.6 &51.8 &91.2 &59.0 &58.4 &41.5 &54.5 &70.0 &64.2\\
        MADAv2 &\textbf{97.4} &\textbf{79.8} &\textbf{90.3} &\textbf{46.6} &\textbf{52.5} &\textbf{54.3} &\textbf{62.3} &\textbf{71.7} &\textbf{91.2} &\textbf{50.2} &\textbf{94.0} &\textbf{77.8} &\textbf{56.8} &\textbf{93.2} &\textbf{74.7} &\textbf{80.3} &\textbf{52.4} &\textbf{57.5} &\textbf{73.6} &\textbf{65.1}\\
    \bottomrule[1pt]
	\end{tabular}}
    }
\end{table*}

\subsection{Ablation Study}
\label{sec:ablation}
To verify the effectiveness of each component, we perform ablation study with the following variants: $\mathbf{M}^{(0)}$: the baseline adversarial learning method \cite{tsai2018learning} without any active annotation; $\mathbf{M}^{(1)}$: extending $\mathbf{M}^{(0)}$ by additionally introducing the active samples with cross entropy loss for training; $\mathbf{M}^{(2)}$: extending $\mathbf{M}^{(1)}$ by adding the cross entropy based prediction consistency loss and introducing the ST strategy; $\mathbf{M}^{(3)}$: replacing the cross entropy loss with the OHEM loss in $\mathbf{M}^{(1)}$; $\mathbf{M}^{(4)}$: extending $\mathbf{M}^{(3)}$ by augmenting target images with ACM and ACP; $\mathbf{M}^{(5)}$: adding the proposed multi-anchor soft alignment loss on target samples in addition to $\mathbf{M}^{(4)}$; 
$\mathbf{M}^{(u)}$: performing fully-supervised segmentation with the annotation of both the source and target datasets as the upper bound.
As shown in Table~\ref{table:ablation_study}, the consistent and notable improvements from $\mathbf{M}^{(0)}$ to $\mathbf{M}^{(5)}$ on two adaptation tasks demonstrate the effectiveness of each strategy. Taking the GTA5 to Cityscapes adaptation as an example, the mIoU rises from 42.5\% to 65.1\% by combining active samples, and reaches 68.5\% with the additional self-traing strategy. The performance continues to improve steadily by 1.0\% after introducing the OHEM loss, and 1.3\% for ACM and ACP, demonstrating that ACM, ACP and OHEM can urge the model focusing more on the
difficult pixels, and effectively address the long-tail issue. Furthermore, the soft-anchor alignment loss leads to a boost of 0.6\% by learning the underlying target-domain distribution. Similar trends can be observed on the SYNTHIA to Cityscapes adaptation task as well.
At last, MADAv2 with only 5\% of the target-domain samples actively annotated achieves comparable performance to that of the upper bound. This demonstrates that the proposed framework can select complementary samples to effectively shrink the performance gap between UDA and full supervision.

The visualization of the feature distribution with/without active learning is presented in Fig.~\ref{fig:cluster_visual_1}. With the proposed MADAv2 framework, the target-specific information can be maintained as its original multimodal distribution.

\subsection{Comparison of Sample Selection Methods}
\label{sec:sample_selection}
The performance of active learning depends heavily on the sample selection methods.
In Table~\ref{table:selection}, we compare the proposed anchor-based method with the following popular sample selection approaches on the GTA5 to Cityscapes adaptation task. 

\noindent
\textbf{Random Selection.} Samples are randomly selected with equal probability from the target domain.

\noindent
\textbf{Entropy-based Uncertainty Method.} The AdvEnt~\cite{vu2019advent} is applied to obtain the prediction map entropy of each sample in the target domain and the ones with top $5\%$ highest entropy are chosen for manual annotation:
\begin{equation}
    {E}_{{ent}}=\frac{-1}{\log (C)} \sum_{c=1}^{C}\sum_{i=1}^{H \times W} p_{i,c}^{t} \log (p_{i,c}^{t}).
\end{equation}

\noindent
\textbf{Adversarial-based Diversity Method.} With the discriminator $f_D$ trained in the warm-up stage as~\cite{tsai2018learning}, we select the samples with least predicted probabilities, \emph{i.e.}, the ones that are the most distinguishable from the source domain:
\begin{equation}
    {E}_{adv}=\frac{1-f_D(f_E(x^t)}{f_D(f_E(x^t))}.
\end{equation}

\noindent
\textbf{AADA Method.} In addition to the discriminator-based diversity, the AADA~\cite{su2020active} method also takes the certainty of prediction into consideration: 
\begin{equation}
    E_{AADA}=E_{ent}E_{adv}.
\end{equation}

\noindent
\textbf{Prototype Method.} To verify the effectiveness of multimodal distributions, we also compare with the unimodal prototype method, \emph{i.e.}, ProDA~\cite{zhang2021prototypical}. If we define $\eta$ as the single centroid,  the samples with a large distance to the centorid are selected:
\begin{equation}
    E_{Proto} = \left\|F^{t}(x^t)-\eta\right\|_{2}^{2}.
\end{equation}

Note that for a fair comparison, all the comparison experiments are subject to the same experimental setup. The same percentage of active samples, $5\%$, are selected, and no unlabeled samples are used for optimization. From the results, we observe that the proposed  \textit{Dual\_Domain\_Distance} strategy delivers the best segmentation performance, demonstrating the superiority of the proposed strategy and the effectiveness of considering the distributions of both domains for sample selection.

\begin{table}[htp]
    \caption{Comparison between MADA and MADAv2.}
    \label{table:cross_ablation}
    \centering
    \setlength\tabcolsep{1.9mm}
    \renewcommand\arraystretch{1.1}
	\def \mysp {\hspace{7pt}}
    {\small 
    \scalebox{1.0}{
	\begin{tabular}{lcc}
    \toprule[1pt]
    \multicolumn{3}{c}{GTA5 $\rightarrow$ Cityscapse}\\
    \toprule
    \diagbox{Semi}{Sample} &MADA &MADAv2 \\
    \toprule
    MADA &66.9 &67.5  \\
    \toprule
    MADAv2 &70.5 &71.4 \\
    \bottomrule[1pt]
    \end{tabular}}
    }
\end{table}

To further evaluate the effect of different sample selection approaches used in MADA and  MADAv2 as well as the semi-supervised learning methods, we perform an ablation study in Table~\ref{table:cross_ablation}, with \emph{Sample} denotes different sample selection approaches and \emph{Semi} represents different semi-supervised learning methods. The experiments show that 1) the sample selection strategy of MADAv2 outperforms the previous MADA on both coarse semi-supervision of MADA (66.9\% to 67.5\% in mIoU) and refined semi-supervision of MADAv2 (70.5\% to 71.4\% in mIoU), 2) the semi-supervised learning method of MADAv2 is also better than the one of MADA regardless of the selected samples.

\subsection{Impact of Source Information}
\label{sec:impact_of_source_info}
As demonstrated in Table~\ref{table:ablation_study}, the performance is improved steadily with ST-based semi-supervised learning. One may wonder how much the source information contributes to the performance, or whether semi-supervised learning with only active samples can achieve similar results.

To investigate this question, we conduct an additional experiment by modifying our MADAv2 from two aspects: 1) only considering the distance from target samples to their nearest target anchor as the selection metric; 2) only training on target labeled or unlabeled samples. The final performance is significantly lower than MADAv2 (67.3\% compared to 71.4\% in mIoU), indicating that the source samples can provide effective information when the amount of target annotations is limited.

\subsection{Impact of the Number of Active Samples}
\label{sec:impact_of_sample_num}
In order to verify the stability of our sample selection metric, comparative experiments regarding different percentages of active samples are conducted. To avoid the effect of unlabeled samples and semi-supervised learning method, only the source samples and labeled target samples are adopted to train the network in a fully supervised manner for this experiment.  
As shown in Table~\ref{table:partion}, as the percentage of samples increases from 1\% to 20\%, the mIoU increases steadily from 57.4\% to 67.3\%.
We also gain the upper bound by optimizing with all target labels, and find a narrow gap of 6.8\% in mIoU  between using only 5\% of target-domain data for AL and the upper bound, demonstrating that the proposed method can effectively exploit the information from active samples. 
We also find a trend that the performance rises sharply by introducing new AL samples when the quantity of samples is small, but improves slowly with further more samples introduced. 
This trend demonstrates that we can claim most of the benefit by selecting a few informative samples at the beginning, and 5\% of samples is a trade-off between annotation cost and segmentation performance.

\begin{table}[!htp]
    \caption{Experiments with different numbers of active samples.}
    \label{table:partion}
    \centering
    \setlength\tabcolsep{1.9mm}
    \renewcommand\arraystretch{1.1}
	\def \mysp {\hspace{7pt}}
    {\small 
    \scalebox{1.0}{
	\begin{tabular}{l@{\ }@{\ } cccccc}
    \toprule[1pt]
    \multicolumn{7}{c}{GTA5 $\rightarrow$ Cityscapse}\\
    \toprule
    Percentage &1\% &2\% &5\% &10\% &20\% &100\% \\
    \toprule
    mIoU &57.4 &59.7 &65.1 &66.5 &67.3 &71.9\\
    \toprule
    mIoU Gap &$-$14.5&$-$12.2&$-$6.8&$-$5.4&$-$4.6 &-\\
    \bottomrule[1pt]
    \end{tabular}}
    }
\end{table}

\subsection{Impact of the Number of Anchors}
\label{sec:impact_of_K}
\begin{figure}[!htp]
	\centering
	\includegraphics[width=1.0\columnwidth]{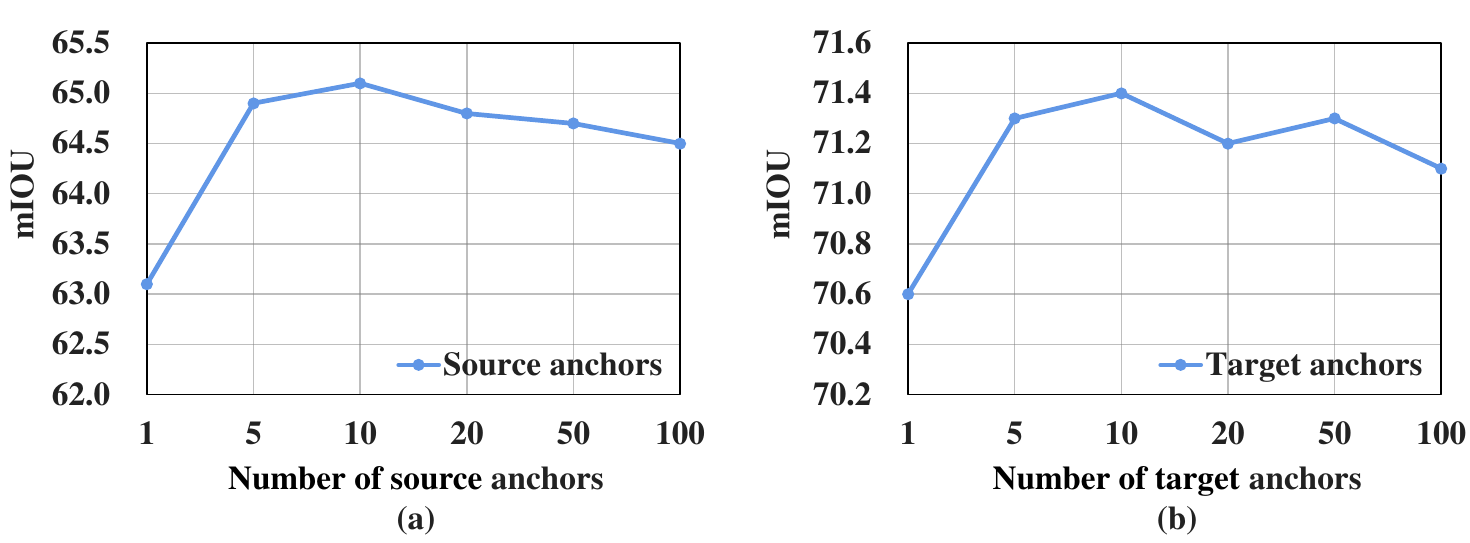}
	\caption{Experiments on different numbers of anchors for the source domain (a) and the target domain (b).}
	\label{fig:n_of_anchors}
\end{figure}

We evaluate the impact of different anchor numbers on modeling the source and target domains with the GTA5 to Cityscapes adaptation task, where the number of anchors varies from 1 to 100 in one domain while fixing the anchor number in the other domain to 10.. As shown in Fig.~\ref{fig:n_of_anchors}, for both domains, using multiple anchors performs consistently better than using a single centroid, and using 5-10 anchors stably yields superior performance.
This might be because there are only limited types of scenarios in these datasets, and a few anchors are sufficient to representing their distributions.
We therefore use 10 clusters considering the top performance in both domains.

\begin{table}[!htp]
    \caption{Result of computational complexity analysis.}
    \label{table:complexity}
    \centering
    \setlength\tabcolsep{1.9mm}
    \renewcommand\arraystretch{1.1}
	\def \mysp {\hspace{7pt}}
    {\small 
    \scalebox{1.0}{
\begin{tabular}{l@{\ }@{\ } ccccc}
    \toprule[1pt]
     &Params &FLOPs &Mem &T{\textunderscore}train &T{\textunderscore}infer \\
    \toprule
    Base &59.3 M &156 G &1.3 GB &77.3 ms &5.6 ms\\
    \toprule
    MADAv2 &118.6 M &312 G &3.2 GB &183.4 ms &5.6 ms\\
    \bottomrule[1pt]
    \end{tabular}}
    }
\end{table}

\subsection{Computational Complexity Analysis}
Here, we present the computational complexity analysis of our method. The \emph{Base} denotes training the DeepLab v3+ model with the cross-entropy loss, while \emph{MADAv2} represents the proposed semi-supervised method. Note that all the results are obtained with single image of 896 $\times$ 512 as input. The parameters, FLOPs and memory are obtained during training. The T{\textunderscore}train and T{\textunderscore}infer denote the time of training and prediction, respectively. The parameters and FLOPs are doubled to the base model due to the requirement of a prediction model. The memory and time cost are greater than twice due to the intermediate variables and extra computation for semi-supervision learning, respectively. However, both methods share the same the inference time because networks with the same architecture (DeepLab v3+) are used for their inference.

\section{Conclusion}
In this paper, we proposed an advanced multi-anchor based
active domain adaptation segmentation framework, namely MADAv2, to reduce the distortion of source-to-target domain adaptation for segmentation tasks at minimal annotation cost.
MADAv2 performs anchor-based active sample selection in DA, which selects only a limited number of target-domain samples that are however most complementary to the source-domain distribution and meanwhile unique to the target-domain distribution.
Adding active annotations of these selected target-domain samples to training can effectively alleviate the distortion of the target-domain distribution that could otherwise happen to typical UDA methods.
Different from previous works which assume unimodal distributions for both the source and target domains, MADAv2 uses multiple anchors to realize multimodal distributions for both domains.
On top of that, MADAv2 further introduces ACM, ACP and OHEM loss to address the long-tail issue. With the multi-anchor soft-alignment loss to explicitly push the target-domain features towards the anchors, the unlabeled target-domain samples can be fully utilized.
Experimental results on two public benchmark datasets have demonstrated the effectiveness of 1) introducing AL into DA, 2) multiple anchors versus a single centroid, 3) introduction of ACM, ACP and OHEM, 4) adding the soft-alignment loss, as well as the superior performance of MADAv2 to existing state-of-the-art UDA and active DA methods.

\section*{Acknowledgement}
This work was supported by National Key R\&D Program of China  2022ZD0118101, Natural Science Foundation of China (No.62202014), Shenzhen Basic Research Program under Grant JCYJ20220813151736001, and National Key R\&D Program of China under Grant 2020AAA0109500/2020AAA0109501, and also sponsored by CCF Tencent Open Research Fund.

\bibliographystyle{IEEEtran}
\bibliography{egbib}

\begin{thebibliography}{10}
\providecommand{\url}[1]{#1}
\csname url@samestyle\endcsname
\providecommand{\newblock}{\relax}
\providecommand{\bibinfo}[2]{#2}
\providecommand{\BIBentrySTDinterwordspacing}{\spaceskip=0pt\relax}
\providecommand{\BIBentryALTinterwordstretchfactor}{4}
\providecommand{\BIBentryALTinterwordspacing}{\spaceskip=\fontdimen2\font plus
\BIBentryALTinterwordstretchfactor\fontdimen3\font minus
  \fontdimen4\font\relax}
\providecommand{\BIBforeignlanguage}[2]{{%
\expandafter\ifx\csname l@#1\endcsname\relax
\typeout{** WARNING: IEEEtran.bst: No hyphenation pattern has been}%
\typeout{** loaded for the language `#1'. Using the pattern for}%
\typeout{** the default language instead.}%
\else
\language=\csname l@#1\endcsname
\fi
#2}}
\providecommand{\BIBdecl}{\relax}
\BIBdecl

\bibitem{geiger2012we}
A.~Geiger, P.~Lenz, and R.~Urtasun, ``Are we ready for autonomous driving?
  {T}he {KITTI} vision benchmark suite,'' in \emph{Proceedings of the IEEE
  Conference on Computer Vision and Pattern Recognition}, 2012, pp. 3354--3361.

\bibitem{cordts2016cityscapes}
M.~Cordts, M.~Omran, S.~Ramos, T.~Rehfeld, M.~Enzweiler, R.~Benenson,
  U.~Franke, S.~Roth, and B.~Schiele, ``The {C}ityscapes dataset for semantic
  urban scene understanding,'' in \emph{Proceedings of the IEEE Conference on
  Computer Vision and Pattern Recognition}, 2016, pp. 3213--3223.

\bibitem{Piao_2019_DMRA}
Y.~Piao, W.~Ji, J.~Li, M.~Zhang, and H.~Lu, ``Depth-induced multi-scale
  recurrent attention network for saliency detection,'' in \emph{Proceedings of
  the IEEE/CVF International Conference on Computer Vision}, 2019, pp.
  7254--7263.

\bibitem{Ji_2021_DCF}
W.~Ji, J.~Li, S.~Yu, M.~Zhang, Y.~Piao, S.~Yao, Q.~Bi, K.~Ma, Y.~Zheng, H.~Lu,
  and L.~Cheng, ``Calibrated {RGB-D} salient object detection,'' in
  \emph{Proceedings of the IEEE Conference on Computer Vision and Pattern
  Recognition}, June 2021, pp. 9471--9481.

\bibitem{Zhang_2019_MoLF}
M.~Zhang, J.~Li, W.~Ji, Y.~Piao, and H.~Lu, ``Memory-oriented decoder for light
  field salient object detection,'' in \emph{Advances in Neural Information
  Processing Systems}, 2019, pp. 896--906.

\bibitem{li2021joint}
J.~Li, W.~Ji, Q.~Bi, C.~Yan, M.~Zhang, Y.~Piao, H.~Lu \emph{et~al.}, ``Joint
  semantic mining for weakly supervised {RGB-D} salient object detection,''
  \emph{Advances in Neural Information Processing Systems}, vol.~34, pp.
  11\,945--11\,959, 2021.

\bibitem{ji2022promoting}
W.~Ji, J.~Li, Q.~Bi, C.~Guo, J.~Liu, and L.~Cheng, ``Promoting saliency from
  depth: Deep unsupervised {{RGB}-D} saliency detection,'' in
  \emph{International Conference on Learning Representations}, 2022.

\bibitem{oberweger2015hands}
M.~Oberweger, P.~Wohlhart, and V.~Lepetit, ``Hands deep in deep learning for
  hand pose estimation,'' \emph{arXiv preprint arXiv:1502.06807}, 2015.

\bibitem{Ji_2021_MRNet}
W.~Ji, S.~Yu, J.~Wu, K.~Ma, C.~Bian, Q.~Bi, J.~Li, H.~Liu, L.~Cheng, and
  Y.~Zheng, ``Learning calibrated medical image segmentation via multi-rater
  agreement modeling,'' in \emph{Proceedings of the IEEE Conference on Computer
  Vision and Pattern Recognition}, 2021, pp. 12\,341--12\,351.

\bibitem{ning2020macro}
M.~Ning, C.~Bian, D.~Lu, H.-Y. Zhou, S.~Yu, C.~Yuan, Y.~Guo, Y.~Wang, K.~Ma,
  and Y.~Zheng, ``A macro-micro weakly-supervised framework for {AS-OCT} tissue
  segmentation,'' in \emph{International Conference on Medical Image Computing
  and Computer-Assisted Intervention}.\hskip 1em plus 0.5em minus 0.4em\relax
  Springer, 2020, pp. 725--734.

\bibitem{ning2021ensembled}
M.~Ning, C.~Bian, C.~Yuan, K.~Ma, and Y.~Zheng, ``Ensembled {R}es{U}net for
  anatomical brain barriers segmentation,'' \emph{Segmentation, Classification,
  and Registration of Multi-modality Medical Imaging Data}, vol. 12587, p.~27,
  2021.

\bibitem{ning2021new}
M.~Ning, C.~Bian, D.~Wei, S.~Yu, C.~Yuan, Y.~Wang, Y.~Guo, K.~Ma, and Y.~Zheng,
  ``A new bidirectional unsupervised domain adaptation segmentation
  framework,'' in \emph{International Conference on Information Processing in
  Medical Imaging}.\hskip 1em plus 0.5em minus 0.4em\relax Springer, 2021, pp.
  492--503.

\bibitem{ma2021abdomenct}
J.~Ma, Y.~Zhang, S.~Gu, C.~Zhu, C.~Ge, Y.~Zhang, X.~An, C.~Wang, Q.~Wang,
  X.~Liu \emph{et~al.}, ``Abdomen{CT-1K}: Is abdominal organ segmentation a
  solved problem,'' \emph{IEEE Transactions on Pattern Analysis and Machine
  Intelligence}, 2021.

\bibitem{choi2020cars}
S.~Choi, J.~T. Kim, and J.~Choo, ``Cars can't fly up in the sky: Improving
  urban-scene segmentation via height-driven attention networks,'' in
  \emph{Proceedings of the IEEE/CVF Conference on Computer Vision and Pattern
  Recognition}, 2020, pp. 9373--9383.

\bibitem{van2008visualizing}
L.~Van~der Maaten and G.~Hinton, ``Visualizing data using t-{SNE}.''
  \emph{Journal of Machine Learning Research}, vol.~9, no.~11, 2008.

\bibitem{tsai2018learning}
Y.-H. Tsai, W.-C. Hung, S.~Schulter, K.~Sohn, M.-H. Yang, and M.~Chandraker,
  ``Learning to adapt structured output space for semantic segmentation,'' in
  \emph{Proceedings of the IEEE Conference on Computer Vision and Pattern
  Recognition}, 2018, pp. 7472--7481.

\bibitem{chen2017no}
Y.-H. Chen, W.-Y. Chen, Y.-T. Chen, B.-C. Tsai, Y.-C. Frank~Wang, and M.~Sun,
  ``No more discrimination: Cross city adaptation of road scene segmenters,''
  in \emph{Proceedings of the IEEE International Conference on Computer
  Vision}, 2017, pp. 1992--2001.

\bibitem{hoffman2018cycada}
J.~Hoffman, E.~Tzeng, T.~Park, J.-Y. Zhu, P.~Isola, K.~Saenko, A.~Efros, and
  T.~Darrell, ``Cycada: Cycle-consistent adversarial domain adaptation,'' in
  \emph{International Conference on Machine Learning}, 2018, pp. 1989--1998.

\bibitem{hoffman2016fcns}
J.~Hoffman, D.~Wang, F.~Yu, and T.~Darrell, ``{FCN}s in the wild: Pixel-level
  adversarial and constraint-based adaptation,'' \emph{arXiv preprint
  arXiv:1612.02649}, 2016.

\bibitem{settles2009active}
B.~Settles, ``Active learning literature survey,'' Department of Computer
  Sciences, University of Wisconsin-Madison, Tech. Rep., 2009.

\bibitem{su2020active}
J.-C. Su, Y.-H. Tsai, K.~Sohn, B.~Liu, S.~Maji, and M.~Chandraker, ``Active
  adversarial domain adaptation,'' in \emph{The IEEE Winter Conference on
  Applications of Computer Vision}, 2020, pp. 739--748.

\bibitem{ning2021multi}
M.~Ning, D.~Lu, D.~Wei, C.~Bian, C.~Yuan, S.~Yu, K.~Ma, and Y.~Zheng,
  ``Multi-anchor active domain adaptation for semantic segmentation,'' in
  \emph{Proceedings of the IEEE/CVF International Conference on Computer
  Vision}, 2021, pp. 9112--9122.

\bibitem{richter2016playing}
S.~R. Richter, V.~Vineet, S.~Roth, and V.~Koltun, ``Playing for data: Ground
  truth from computer games,'' in \emph{European Conference on Computer
  Vision}.\hskip 1em plus 0.5em minus 0.4em\relax Springer, 2016, pp. 102--118.

\bibitem{ros2016synthia}
G.~Ros, L.~Sellart, J.~Materzynska, D.~Vazquez, and A.~M. Lopez, ``The
  {S}ynthia dataset: A large collection of synthetic images for semantic
  segmentation of urban scenes,'' in \emph{Proceedings of the IEEE Conference
  on Computer Vision and Pattern Recognition}, 2016, pp. 3234--3243.

\bibitem{hu2021semi}
H.~Hu, F.~Wei, H.~Hu, Q.~Ye, J.~Cui, and L.~Wang, ``Semi-supervised semantic
  segmentation via adaptive equalization learning,'' \emph{Advances in Neural
  Information Processing Systems}, vol.~34, pp. 22\,106--22\,118, 2021.

\bibitem{shrivastava2016training}
A.~Shrivastava, A.~Gupta, and R.~Girshick, ``Training region-based object
  detectors with online hard example mining,'' in \emph{Proceedings of the IEEE
  Conference on Computer Vision and Pattern Recognition}, 2016, pp. 761--769.

\bibitem{glorot2011domain}
X.~Glorot, A.~Bordes, and Y.~Bengio, ``Domain adaptation for large-scale
  sentiment classification: A deep learning approach,'' in \emph{International
  Conference on Machine Learning}, 2011.

\bibitem{li2019joint}
S.~Li, C.~H. Liu, B.~Xie, L.~Su, Z.~Ding, and G.~Huang, ``Joint adversarial
  domain adaptation,'' in \emph{Proceedings of the 27th ACM International
  Conference on Multimedia}, 2019, pp. 729--737.

\bibitem{li2021transferable}
S.~Li, M.~Xie, K.~Gong, C.~H. Liu, Y.~Wang, and W.~Li, ``Transferable semantic
  augmentation for domain adaptation,'' in \emph{Proceedings of the IEEE/CVF
  Conference on Computer Vision and Pattern Recognition}, 2021, pp.
  11\,516--11\,525.

\bibitem{long2018conditional}
M.~Long, Z.~Cao, J.~Wang, and M.~I. Jordan, ``Conditional adversarial domain
  adaptation,'' \emph{Advances in Neural Information Processing Systems},
  vol.~31, 2018.

\bibitem{chen2018domain}
Y.~Chen, W.~Li, C.~Sakaridis, D.~Dai, and L.~Van~Gool, ``Domain adaptive faster
  {R-CNN} for object detection in the wild,'' in \emph{Proceedings of the IEEE
  Conference on Computer Vision and Pattern Recognition}, 2018, pp. 3339--3348.

\bibitem{vs2021mega}
V.~Vs, V.~Gupta, P.~Oza, V.~A. Sindagi, and V.~M. Patel, ``Mega-{CDA}: Memory
  guided attention for category-aware unsupervised domain adaptive object
  detection,'' in \emph{Proceedings of the IEEE/CVF Conference on Computer
  Vision and Pattern Recognition}, 2021, pp. 4516--4526.

\bibitem{liu2020open}
Z.~Liu, Z.~Miao, X.~Pan, X.~Zhan, D.~Lin, S.~X. Yu, and B.~Gong, ``Open
  compound domain adaptation,'' in \emph{Proceedings of the IEEE/CVF Conference
  on Computer Vision and Pattern Recognition}, 2020, pp. 12\,406--12\,415.

\bibitem{long2015learning}
M.~Long, Y.~Cao, J.~Wang, and M.~Jordan, ``Learning transferable features with
  deep adaptation networks,'' in \emph{International Conference on Machine
  Learning}, 2015, pp. 97--105.

\bibitem{long2017deep}
M.~Long, H.~Zhu, J.~Wang, and M.~I. Jordan, ``Deep transfer learning with joint
  adaptation networks,'' in \emph{International Conference on Machine
  Learning}, 2017, pp. 2208--2217.

\bibitem{zellinger2017central}
W.~Zellinger, T.~Grubinger, E.~Lughofer, T.~Natschl{\"a}ger, and
  S.~Saminger-Platz, ``{Central moment discrepancy (CMD) for domain-invariant
  representation learning},'' \emph{arXiv preprint arXiv:1702.08811}, 2017.

\bibitem{sun2016deep}
B.~Sun and K.~Saenko, ``Deep {CORAL}: Correlation alignment for deep domain
  adaptation,'' in \emph{European Conference on Computer Vision}.\hskip 1em
  plus 0.5em minus 0.4em\relax Springer, 2016, pp. 443--450.

\bibitem{zhu2017unpaired}
J.-Y. Zhu, T.~Park, P.~Isola, and A.~A. Efros, ``Unpaired image-to-image
  translation using cycle-consistent adversarial networks,'' in
  \emph{Proceedings of the IEEE International Conference on Computer Vision},
  2017, pp. 2223--2232.

\bibitem{li2019bidirectional}
Y.~Li, L.~Yuan, and N.~Vasconcelos, ``Bidirectional learning for domain
  adaptation of semantic segmentation,'' in \emph{Proceedings of the IEEE
  Conference on Computer Vision and Pattern Recognition}, 2019, pp. 6936--6945.

\bibitem{chang2019all}
W.-L. Chang, H.-P. Wang, W.-H. Peng, and W.-C. Chiu, ``All about structure:
  Adapting structural information across domains for boosting semantic
  segmentation,'' in \emph{Proceedings of the IEEE Conference on Computer
  Vision and Pattern Recognition}, 2019, pp. 1900--1909.

\bibitem{huang2018multimodal}
X.~Huang, M.-Y. Liu, S.~Belongie, and J.~Kautz, ``Multimodal unsupervised
  image-to-image translation,'' in \emph{Proceedings of the European Conference
  on Computer Vision}, 2018, pp. 172--189.

\bibitem{luo2019taking}
Y.~Luo, L.~Zheng, T.~Guan, J.~Yu, and Y.~Yang, ``Taking a closer look at domain
  shift: Category-level adversaries for semantics consistent domain
  adaptation,'' in \emph{Proceedings of the IEEE Conference on Computer Vision
  and Pattern Recognition}, 2019, pp. 2507--2516.

\bibitem{zhang2019category}
Q.~Zhang, J.~Zhang, W.~Liu, and D.~Tao, ``Category anchor-guided unsupervised
  domain adaptation for semantic segmentation,'' in \emph{Advances in Neural
  Information Processing Systems}, 2019, pp. 435--445.

\bibitem{vu2019advent}
T.-H. Vu, H.~Jain, M.~Bucher, M.~Cord, and P.~P{\'e}rez, ``Advent: Adversarial
  entropy minimization for domain adaptation in semantic segmentation,'' in
  \emph{Proceedings of the IEEE Conference on Computer Vision and Pattern
  Recognition}, 2019, pp. 2517--2526.

\bibitem{mei2020instance}
K.~Mei, C.~Zhu, J.~Zou, and S.~Zhang, ``Instance adaptive self-training for
  unsupervised domain adaptation,'' in \emph{European Conference on Computer
  Vision}.\hskip 1em plus 0.5em minus 0.4em\relax Springer, 2020, pp. 415--430.

\bibitem{tranheden2021dacs}
W.~Tranheden, V.~Olsson, J.~Pinto, and L.~Svensson, ``{DACS}: Domain adaptation
  via cross-domain mixed sampling,'' in \emph{Proceedings of the IEEE/CVF
  Winter Conference on Applications of Computer Vision}, 2021, pp. 1379--1389.

\bibitem{jiang2022prototypical}
Z.~Jiang, Y.~Li, C.~Yang, P.~Gao, Y.~Wang, Y.~Tai, and C.~Wang, ``Prototypical
  contrast adaptation for domain adaptive semantic segmentation,'' \emph{arXiv
  preprint arXiv:2207.06654}, 2022.

\bibitem{zhang2021prototypical}
P.~Zhang, B.~Zhang, T.~Zhang, D.~Chen, Y.~Wang, and F.~Wen, ``Prototypical
  pseudo label denoising and target structure learning for domain adaptive
  semantic segmentation,'' in \emph{Proceedings of the IEEE/CVF Conference on
  Computer Vision and Pattern Recognition}, 2021, pp. 12\,414--12\,424.

\bibitem{lu2022bidirectional}
Y.~Lu, Y.~Luo, L.~Zhang, Z.~Li, Y.~Yang, and J.~Xiao, ``Bidirectional
  self-training with multiple anisotropic prototypes for domain adaptive
  semantic segmentation,'' \emph{arXiv preprint arXiv:2204.07730}, 2022.

\bibitem{liu2021swin}
Z.~Liu, Y.~Lin, Y.~Cao, H.~Hu, Y.~Wei, Z.~Zhang, S.~Lin, and B.~Guo, ``Swin
  {T}ransformer: Hierarchical vision {T}ransformer using shifted windows,'' in
  \emph{Proceedings of the IEEE/CVF International Conference on Computer
  Vision}, 2021, pp. 10\,012--10\,022.

\bibitem{hoyer2022hrda}
L.~Hoyer, D.~Dai, and L.~Van~Gool, ``{HRDA}: Context-aware high-resolution
  domain-adaptive semantic segmentation,'' \emph{arXiv preprint
  arXiv:2204.13132}, 2022.

\bibitem{hoyer2022daformer}
------, ``Daformer: Improving network architectures and training strategies for
  domain-adaptive semantic segmentation,'' in \emph{Proceedings of the IEEE/CVF
  Conference on Computer Vision and Pattern Recognition}, 2022, pp. 9924--9935.

\bibitem{cohn1996active}
D.~A. Cohn, Z.~Ghahramani, and M.~I. Jordan, ``Active learning with statistical
  models,'' \emph{Journal of Artificial Intelligence Research}, vol.~4, pp.
  129--145, 1996.

\bibitem{lewis1994heterogeneous}
D.~D. Lewis and J.~Catlett, ``Heterogeneous uncertainty sampling for supervised
  learning,'' in \emph{Proceedings of the International Conference on Machine
  Learning}.\hskip 1em plus 0.5em minus 0.4em\relax Elsevier, 1994, pp.
  148--156.

\bibitem{scheffer2001active}
T.~Scheffer, C.~Decomain, and S.~Wrobel, ``Active hidden {M}arkov models for
  information extraction,'' in \emph{International Symposium on Intelligent
  Data Analysis}.\hskip 1em plus 0.5em minus 0.4em\relax Springer, 2001, pp.
  309--318.

\bibitem{dutt2016active}
S.~Dutt~Jain and K.~Grauman, ``Active image segmentation propagation,'' in
  \emph{Proceedings of the IEEE Conference on Computer Vision and Pattern
  Recognition}, 2016, pp. 2864--2873.

\bibitem{hoi2009semisupervised}
S.~C. Hoi, R.~Jin, J.~Zhu, and M.~R. Lyu, ``Semisupervised {SVM} batch mode
  active learning with applications to image retrieval,'' \emph{ACM
  Transactions on Information Systems}, vol.~27, no.~3, pp. 1--29, 2009.

\bibitem{huang2010active}
S.-J. Huang, R.~Jin, and Z.-H. Zhou, ``Active learning by querying informative
  and representative examples,'' \emph{Advances in Neural Information
  Processing Systems}, vol.~23, pp. 892--900, 2010.

\bibitem{dasgupta2008hierarchical}
S.~Dasgupta and D.~Hsu, ``Hierarchical sampling for active learning,'' in
  \emph{Proceedings of the 25th International Conference on Machine Learning},
  2008, pp. 208--215.

\bibitem{nguyen2004active}
H.~T. Nguyen and A.~Smeulders, ``Active learning using pre-clustering,'' in
  \emph{Proceedings of the twenty-first International Conference on Machine
  Learning}, 2004, p.~79.

\bibitem{freytag2014selecting}
A.~Freytag, E.~Rodner, and J.~Denzler, ``Selecting influential examples: Active
  learning with expected model output changes,'' in \emph{European Conference
  on Computer Vision}.\hskip 1em plus 0.5em minus 0.4em\relax Springer, 2014,
  pp. 562--577.

\bibitem{kading2015active}
C.~Kading, A.~Freytag, E.~Rodner, P.~Bodesheim, and J.~Denzler, ``Active
  learning and discovery of object categories in the presence of unnameable
  instances,'' in \emph{Proceedings of the IEEE Conference on Computer Vision
  and Pattern Recognition}, 2015, pp. 4343--4352.

\bibitem{vezhnevets2012weakly}
A.~Vezhnevets, V.~Ferrari, and J.~M. Buhmann, ``Weakly supervised structured
  output learning for semantic segmentation,'' in \emph{2012 IEEE Conference on
  Computer Vision and Pattern Recognition}.\hskip 1em plus 0.5em minus
  0.4em\relax IEEE, 2012, pp. 845--852.

\bibitem{qi2008two}
G.-J. Qi, X.-S. Hua, Y.~Rui, J.~Tang, and H.-J. Zhang, ``Two-dimensional active
  learning for image classification,'' in \emph{IEEE Conference on Computer
  Vision and Pattern Recognition}.\hskip 1em plus 0.5em minus 0.4em\relax IEEE,
  2008, pp. 1--8.

\bibitem{kao2018localization}
C.-C. Kao, T.-Y. Lee, P.~Sen, and M.-Y. Liu, ``Localization-aware active
  learning for object detection,'' in \emph{Asian Conference on Computer
  Vision}.\hskip 1em plus 0.5em minus 0.4em\relax Springer, 2018, pp. 506--522.

\bibitem{Ji_2020_CoNet}
W.~Ji, J.~Li, M.~Zhang, Y.~Piao, and H.~Lu, ``Accurate {{RGB}-D} salient object
  detection via collaborative learning,'' in \emph{Proceedings of the European
  Conference on Computer Vision}, 2020, pp. 52--69.

\bibitem{Zhang_2020_LFNet}
M.~Zhang, W.~Ji, Y.~Piao, J.~Li, Y.~Zhang, S.~Xu, and H.~Lu, ``{LFNet}: Light
  field fusion network for salient object detection,'' \emph{IEEE Transactions
  on Image Processing}, vol.~29, pp. 6276--6287, 2020.

\bibitem{sun2015active}
Q.~Sun, A.~Laddha, and D.~Batra, ``Active learning for structured probabilistic
  models with histogram approximation,'' in \emph{Proceedings of the IEEE
  Conference on Computer Vision and Pattern Recognition}, 2015, pp. 3612--3621.

\bibitem{chattopadhyay2013joint}
R.~Chattopadhyay, W.~Fan, I.~Davidson, S.~Panchanathan, and J.~Ye, ``Joint
  transfer and batch-mode active learning,'' in \emph{International Conference
  on Machine Learning}, 2013, pp. 253--261.

\bibitem{huang2018cost}
S.-J. Huang, J.-W. Zhao, and Z.-Y. Liu, ``Cost-effective training of deep
  {CNN}s with active model adaptation,'' in \emph{Proceedings of the 24th ACM
  SIGKDD International Conference on Knowledge Discovery \& Data Mining}, 2018,
  pp. 1580--1588.

\bibitem{ganin2016domain}
Y.~Ganin, E.~Ustinova, H.~Ajakan, P.~Germain, H.~Larochelle, F.~Laviolette,
  M.~Marchand, and V.~Lempitsky, ``Domain-adversarial training of neural
  networks,'' \emph{The Journal of Machine Learning Research}, vol.~17, no.~1,
  pp. 2096--2030, 2016.

\bibitem{shin2021labor}
I.~Shin, D.-J. Kim, J.~W. Cho, S.~Woo, K.~Park, and I.~S. Kweon, ``Lab{OR}:
  Labeling only if required for domain adaptive semantic segmentation,'' in
  \emph{Proceedings of the IEEE/CVF International Conference on Computer
  Vision}, 2021, pp. 8588--8598.

\bibitem{you2022pixel}
F.~You, J.~Li, Z.~Chen, and L.~Zhu, ``Pixel exclusion: Uncertainty-aware
  boundary discovery for active cross-domain semantic segmentation,'' in
  \emph{Proceedings of the 30th ACM International Conference on Multimedia},
  2022, pp. 1866--1874.

\bibitem{wu2022d}
T.-H. Wu, Y.-S. Liou, S.-J. Yuan, H.-Y. Lee, T.-I. Chen, K.-C. Huang, and W.~H.
  Hsu, ``D 2 ada: Dynamic density-aware active domain adaptation for semantic
  segmentation,'' in \emph{Computer Vision--ECCV 2022: 17th European
  Conference, Tel Aviv, Israel, October 23--27, 2022, Proceedings, Part
  XXIX}.\hskip 1em plus 0.5em minus 0.4em\relax Springer, 2022, pp. 449--467.

\bibitem{xie2022towards}
B.~Xie, L.~Yuan, S.~Li, C.~H. Liu, and X.~Cheng, ``Towards fewer annotations:
  Active learning via region impurity and prediction uncertainty for domain
  adaptive semantic segmentation,'' in \emph{Proceedings of the IEEE/CVF
  Conference on Computer Vision and Pattern Recognition}, 2022, pp. 8068--8078.

\bibitem{cui2020unified}
H.~Cui, D.~Wei, K.~Ma, S.~Gu, and Y.~Zheng, ``A unified framework for
  generalized low-shot medical image segmentation with scarce data,''
  \emph{IEEE Transactions on Medical Imaging}, 2020.

\bibitem{macqueen1967some}
J.~MacQueen, ``Some methods for classification and analysis of multivariate
  observations,'' in \emph{Proceedings of the Fifth Berkeley Symposium on
  Mathematical Statistics and Probability}, vol.~1, 1967, pp. 281--297.

\bibitem{siddiqui2020viewal}
Y.~Siddiqui, J.~Valentin, and M.~Nie{\ss}ner, ``View{AL}: Active learning with
  viewpoint entropy for semantic segmentation,'' in \emph{Proceedings of the
  IEEE/CVF Conference on Computer Vision and Pattern Recognition}, 2020, pp.
  9433--9443.

\bibitem{xie2016unsupervised}
J.~Xie, R.~Girshick, and A.~Farhadi, ``Unsupervised deep embedding for
  clustering analysis,'' in \emph{International Conference on Machine
  Learning}, 2016, pp. 478--487.

\bibitem{tarvainen2017mean}
A.~Tarvainen and H.~Valpola, ``Mean teachers are better role models:
  Weight-averaged consistency targets improve semi-supervised deep learning
  results,'' \emph{arXiv preprint arXiv:1703.01780}, 2017.

\bibitem{wang2022semi}
Y.~Wang, H.~Wang, Y.~Shen, J.~Fei, W.~Li, G.~Jin, L.~Wu, R.~Zhao, and X.~Le,
  ``Semi-supervised semantic segmentation using unreliable pseudo-labels,'' in
  \emph{Proceedings of the IEEE/CVF Conference on Computer Vision and Pattern
  Recognition}, 2022, pp. 4248--4257.

\bibitem{everingham2015pascal}
M.~Everingham, S.~A. Eslami, L.~Van~Gool, C.~K. Williams, J.~Winn, and
  A.~Zisserman, ``The {PASCAL} visual object classes challenge: A
  retrospective,'' \emph{International Journal of Computer Vision}, vol. 111,
  no.~1, pp. 98--136, 2015.

\bibitem{chen2018encoder}
L.-C. Chen, Y.~Zhu, G.~Papandreou, F.~Schroff, and H.~Adam, ``Encoder-decoder
  with atrous separable convolution for semantic image segmentation,'' in
  \emph{Proceedings of the European Conference on Computer Vision}, 2018, pp.
  801--818.

\bibitem{he2016deep}
K.~He, X.~Zhang, S.~Ren, and J.~Sun, ``Deep residual learning for image
  recognition,'' in \emph{Proceedings of the IEEE Conference on Computer Vision
  and Pattern Recognition}, 2016, pp. 770--778.

\bibitem{deng2009imagenet}
J.~Deng, W.~Dong, R.~Socher, L.-J. Li, K.~Li, and L.~Fei-Fei, ``Image{N}et: A
  large-scale hierarchical image database,'' in \emph{IEEE Conference on
  Computer Vision and Pattern Recognition}, 2009, pp. 248--255.

\bibitem{maas2013rectifier}
A.~L. Maas, A.~Y. Hannun, and A.~Y. Ng, ``Rectifier nonlinearities improve
  neural network acoustic models,'' in \emph{Proceedings of the International
  Conference on Machine Learning}, vol.~30, no.~1, 2013, p.~3.

\end{thebibliography}

\vspace{-.4in}
\begin{IEEEbiography}[{\includegraphics[width=0.85in,keepaspectratio]{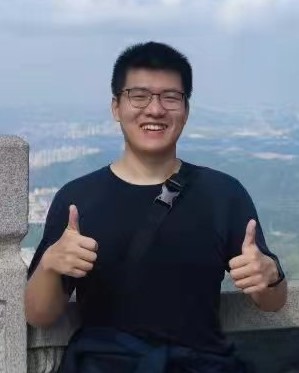}}]
{Munan Ning} is a Ph.D. student currently enrolled at Peking University. He completed his undergraduate studies at Wuhan University and received his Master's degree from the National University of Defense Technology. He has published several papers in top conferences on computer vision and medical imaging. His research interests primarily focus on semi-supervised and domain adaptation learning, and he looks forward to exploring new directions in multimodal and 3D vision.
\end{IEEEbiography}

\vspace{-.4in}
\begin{IEEEbiography}[{\includegraphics[width=0.85in,keepaspectratio]{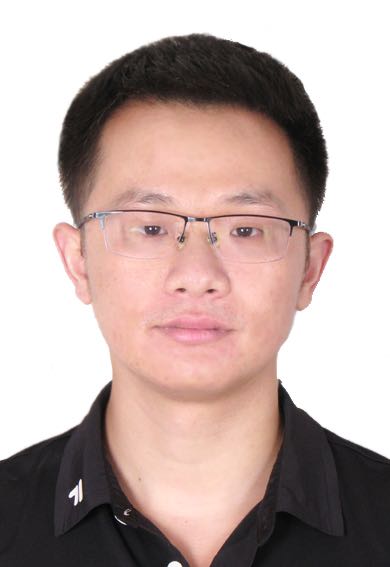}}]
{Donghuan Lu} received his B.E. and M.E. degree in automation science and electrical engineering from Beihang University, China, and the PhD degree in the applied science of Simon Farser University. Now he is a senior researcher at Tencent Jarvis Lab, Shenzhen. His research interests include deep learning and medical image analysis.
\end{IEEEbiography}

\vspace{-.4in}
\begin{IEEEbiography}[{\includegraphics[width=0.85in,keepaspectratio]{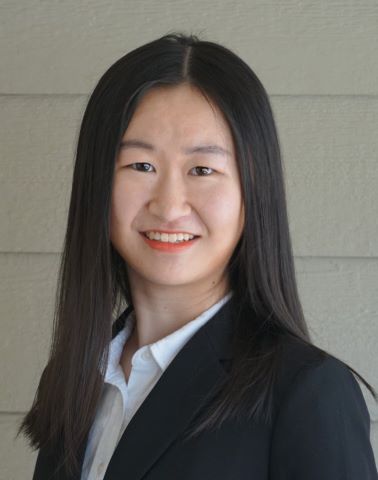}}]
{Yujia Xie} is a researcher at Microsoft, specializing in large-scale training of multimodal representation models. She holds a BS degree from the University of Science and Technology of China and a PhD from Georgia Institute of Technology. Her expertise spans multi-modal learning, language generation, computer vision, and core machine learning, with contributions to vision-language models, GPT grounding, and optimal transport.
\end{IEEEbiography}

\vspace{-.4in}
\begin{IEEEbiography}[{\includegraphics[width=0.85in,keepaspectratio]{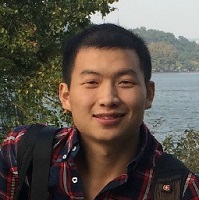}}]
{Dongdong Chen} is a Principal Researcher from Microsoft Research. He received his Ph.D. degree under the joint phd program between University of Science and Technology of China and MSRA. His research interests mainly include generative mode, large-scale pretraining, and general representation learning.
\end{IEEEbiography}

\vspace{-.4in}
\begin{IEEEbiography}[{\includegraphics[width=0.85in,keepaspectratio]{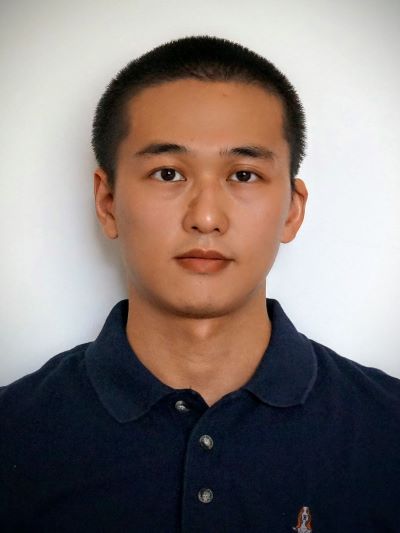}}]
{Dong Wei} received the Ph.D. degree in Computer Engineering from the University of Singapore, Singapore, in 2013. Since 2018, he has been a Senior Researcher at the Tencent Jarvis Lab, Shenzhen, China. His research interests include medical image analysis, with a current focus on data and annotation efficient approaches.
\end{IEEEbiography}

\vspace{-.4in}
\begin{IEEEbiography}[{\includegraphics[width=0.85in,keepaspectratio]{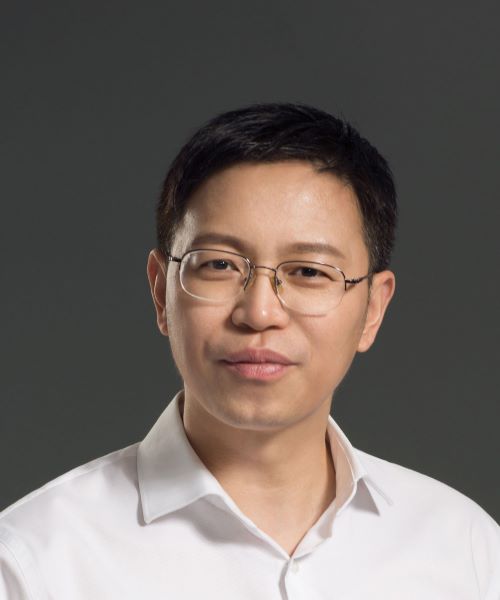}}]
{Yefeng Zheng} (Fellow, IEEE) received the B.E. and M.E. degrees from Tsinghua University, Beijing, in 1998 and 2001, respectively, and the Ph.D. degree from the University of Maryland, College Park, MD, USA, in 2005. After graduation, he joined Siemens Corporate Research, Princeton, NJ, USA. He is currently the Director and the Distinguished Scientist of Tencent Jarvis Lab, Shenzhen, China, leading the company’s initiative on Medical AI. His research interests include medical image analysis, computer vision, natural language processing, and deep learning. Dr. Zheng is a fellow of the American Institute for Medical and Biological Engineering (AIMBE).
\end{IEEEbiography}

\begin{IEEEbiography}[{\includegraphics[width=0.85in,keepaspectratio]{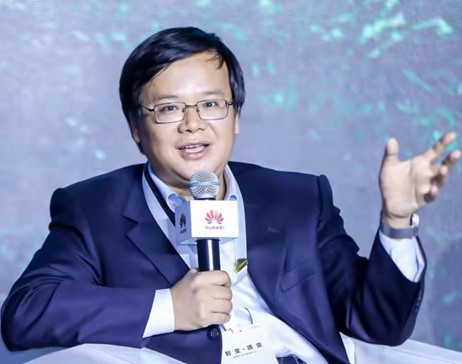}}]
{Yonghong Tian} is the Dean of the School of Electronics and Computer Engineering and a Boya Distinguished Professor at Peking University. He is also the deputy director of Artificial Intelligence Research at PengCheng Laboratory, Shenzhen. His research interests include neuromorphic vision, distributed machine learning, and multimedia big data. He has authored or coauthored over 300 articles in technical journals and conferences. Prof. Tian is a Fellow of IEEE, a senior member of CIE and CCF, and a member of ACM. He has received multiple awards, including the Chinese National Science Foundation for Distinguished Young Scholars, the 2015 EURASIP Best Paper Award, and the 2022 IEEE SA Standards Medallion and SA Emerging Technology Award.
\end{IEEEbiography}

\begin{IEEEbiography}[{\includegraphics[width=1.1in,keepaspectratio]{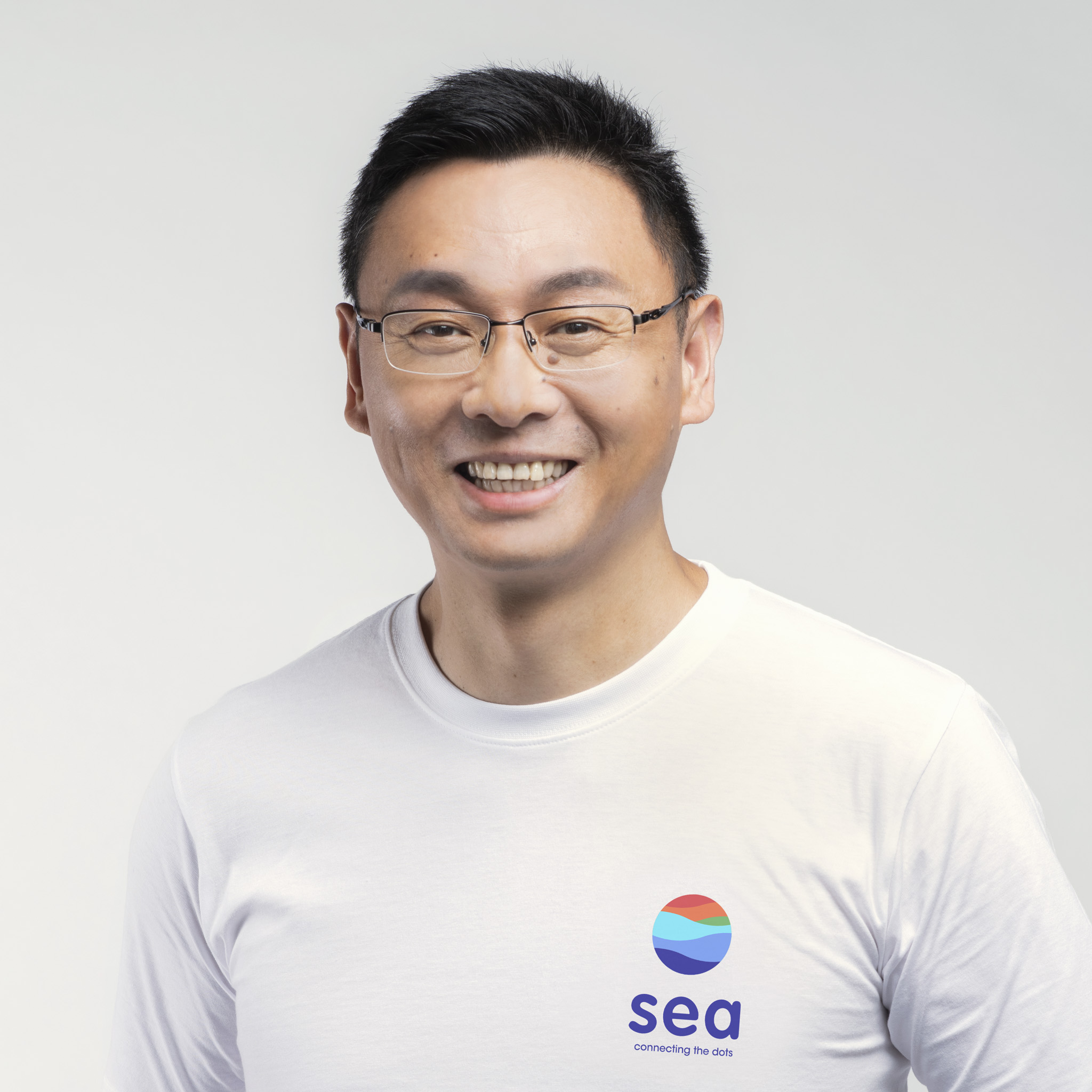}}]
{Shuicheng Yan} is currently director of Sea AI Lab (SAIL) and group chief scientist of Sea. He is a Fellow of Academy of Engineering, Singapore, AAAI Fellow, ACM Fellow, IEEE Fellow, IAPR Fellow. His research areas include computer vision, machine learning and multimedia analysis. Till now, he has published over 600 papers in top international journals and conferences, with H-index 120+. He had been among “Thomson Reuters Highly Cited Researchers” in 2014, 2015, 2016, 2018, 2019, 2020 and 2021.Prof. Yan's team has received winner or honorable-mention prizes for 10 times of two core competitions, Pascal VOC and ImageNet (ILSVRC), which are deemed as “World Cup” in the computer vision community. Also his team won over 10 best paper or best student paper prizes and especially, a grand slam in ACM MM, the top conference in multimedia, including Best Paper Award three times, Best Student Paper Award twice and Best Demo Award once.
\end{IEEEbiography}

\begin{IEEEbiography}[{\includegraphics[width=0.85in,keepaspectratio]{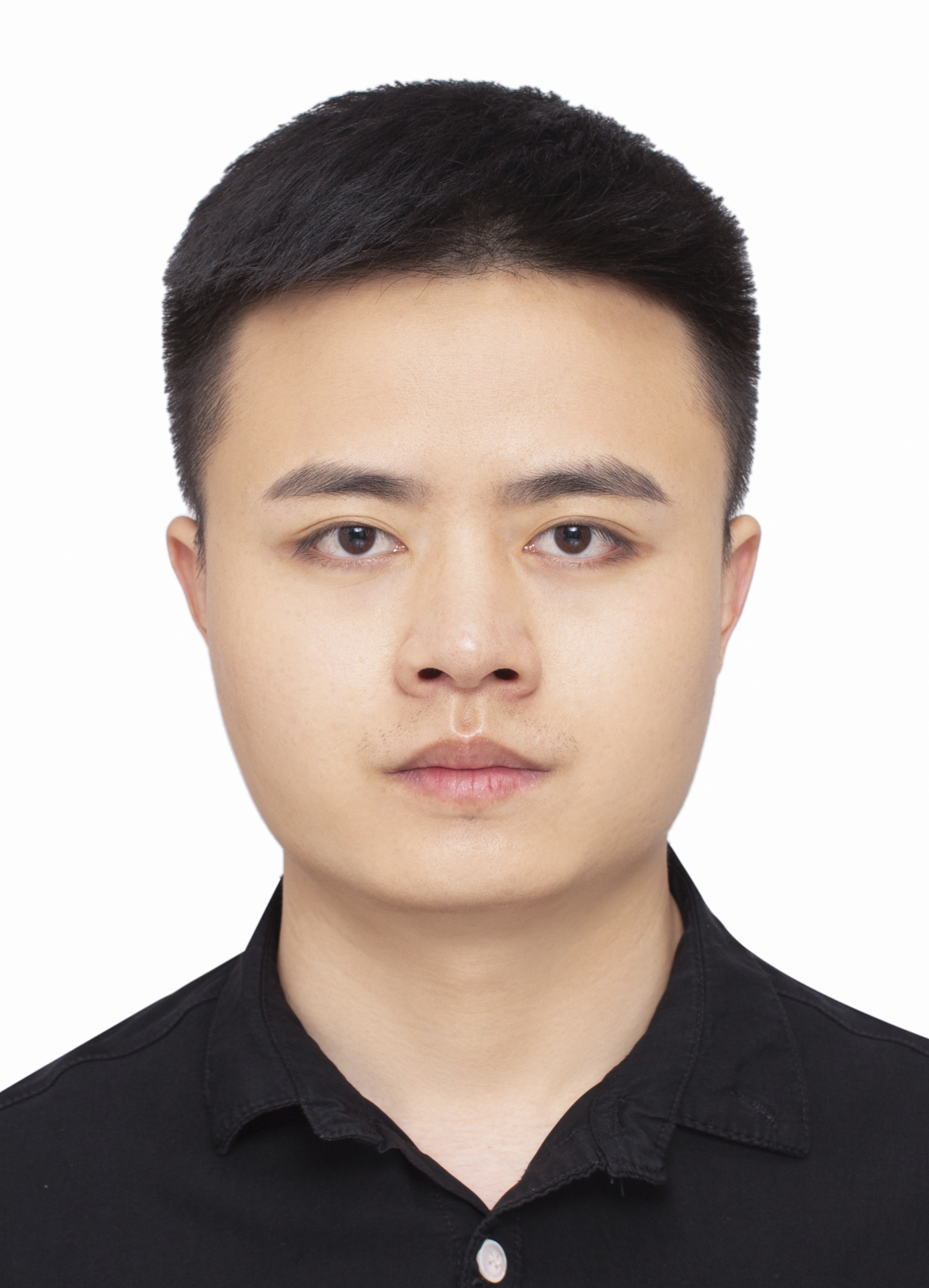}}]
{Li Yuan} received the B.Eng. degree from University of Science and Technology of China in 2017, and Ph.D. degree from National University of Singapore in 2021. He is currently a tenure-track Assistant Professor with School of Electrical and Computer Engineering at Peking University. He has published more than 40 papers on top conferences/journals. His research interests include deep learning, image processing, and computer vision.
\end{IEEEbiography}

\vfill

\end{document}